\documentclass[journal]{IEEEtran}
\usepackage{amsmath,amsfonts}
\usepackage{algorithmic}
\usepackage{algorithm}
\usepackage{array}
\usepackage[caption=false,font=normalsize,labelfont=sf,textfont=sf]{subfig}
\usepackage{textcomp}
\usepackage{stfloats}
\usepackage{url}
\usepackage{verbatim}
\usepackage{graphicx}
\usepackage{cite}
\usepackage{multirow}
\usepackage{booktabs}
\usepackage{xcolor}
\begin{document}

\title{Toward Generalizable Forgery Detection and Reasoning}

\author{Yueying Gao,~Dongliang Chang,~Bingyao Yu,~Haotian Qin,~Muxi Diao,\\~Lei Chen,~\IEEEmembership{Member,~IEEE},~Kongming Liang,
and Zhanyu~Ma,~\IEEEmembership{Senior Member,~IEEE}
\thanks{Y. Gao, D. Chang, H. Qin, M. Diao, K. Lang, and Z. Ma are with the Pattern Recognition and Intelligent
System Laboratory, School of Artificial Intelligence, Beijing University of Posts and Telecommunications, Beijing 100876, China. E-mail: \{gaoyueying, changdongliang, qinhaotian, dmx, liangkongming, mazhanyu\}@bupt.edu.cn}
\thanks{B. Yu and L. Chen are with the Department of Automation, Tsinghua University, Beijing 100084, China. E-mail: \{yuby, leichenthu\}@tsinghua.edu.cn}
\thanks{Corresponding author: Dongliang Chang.}}

\markboth{Journal of \LaTeX\ Class Files,~Vol.~14, No.~8, August~2021}%
{Shell \MakeLowercase{\textit{et al.}}: A Sample Article Using IEEEtran.cls for IEEE Journals}

\IEEEpubid{0000--0000/00\$00.00~\copyright~2021 IEEE}

\maketitle
\begin{abstract}
Accurate and interpretable detection of AI-generated images is essential for mitigating risks associated with AI misuse. However, the substantial domain gap among generative models makes it challenging to develop a generalizable forgery detection model. Moreover, since every pixel in an AI-generated image is synthesized, traditional saliency-based forgery explanation methods are not well suited for this task. To address these challenges, we formulate detection and explanation as a unified Forgery Detection and Reasoning task (\textbf{FDR-Task})\footnote{Throughout this paper, ``forgery'' specifically refers to fully AI-generated images, rather than traditional partial manipulations.}, leveraging Multi-Modal Large Language Models (MLLMs) to provide accurate detection through reliable reasoning over forgery attributes. To facilitate this task, we introduce the Multi-Modal Forgery Reasoning dataset (\textbf{MMFR-Dataset}), a large-scale dataset containing 120K images across 10 generative models, with 378K reasoning annotations on forgery attributes, enabling comprehensive evaluation of the FDR-Task. Furthermore, we propose \textbf{FakeReasoning}, a forgery detection and reasoning framework with three key components: 1) a dual-branch visual encoder that integrates CLIP and DINO to capture both high-level semantics and low-level artifacts; 2) a Forgery-Aware Feature Fusion Module that leverages DINO's attention maps and cross-attention mechanisms to guide MLLMs toward forgery-related clues; 3) a Classification Probability Mapper that couples language modeling and forgery detection, enhancing overall performance. Experiments across multiple generative models demonstrate that FakeReasoning not only achieves robust generalization but also outperforms state-of-the-art methods on both detection and reasoning tasks. The code is available at: https://github.com/PRIS-CV/FakeReasoning.
\end{abstract}

\begin{figure*}
     \centering
     \includegraphics[width=\textwidth]{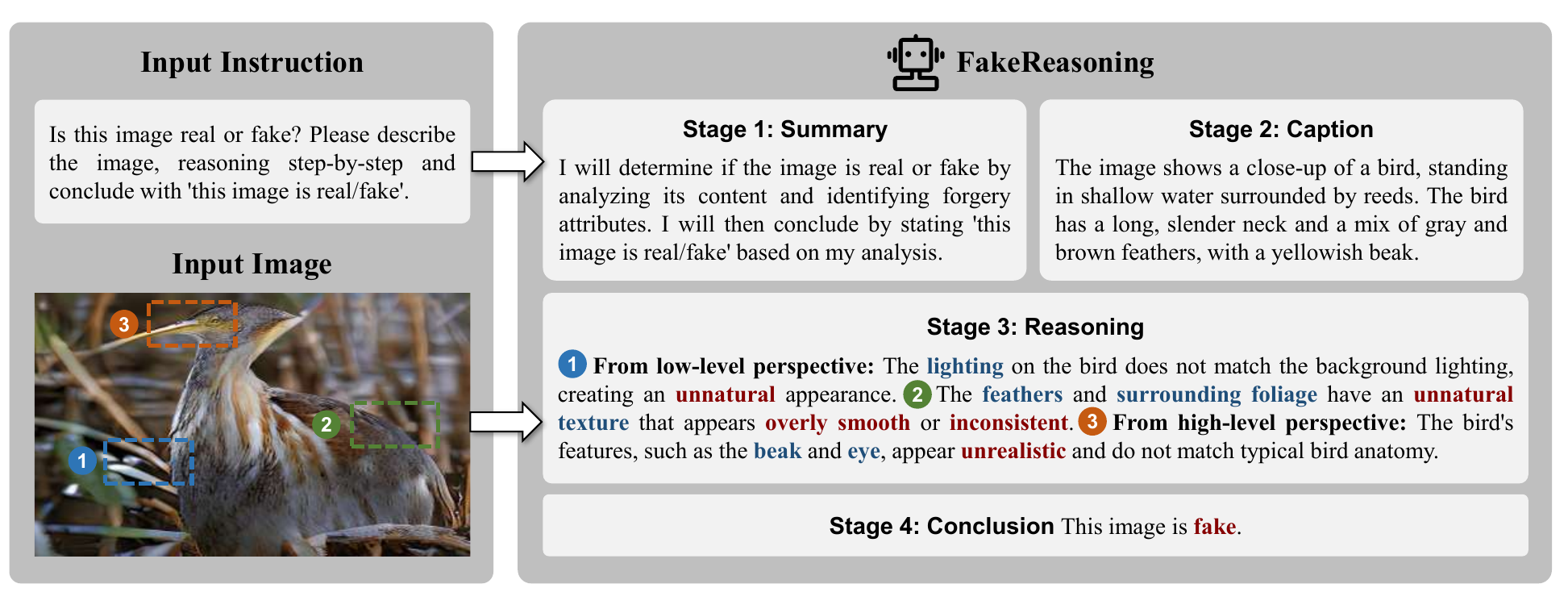}
     \caption{Illustration of the FDR-Task. Different from traditional forgery detection, the FDR-Task leverages MLLMs to perform accurate detection
    through reliable reasoning over forgery attributes, improving both detection accuracy and interpretability.}
     \label{fig:FDR-Task}
\end{figure*}

\begin{IEEEkeywords}
Forgery Detection, Multi-Modal Large Language Model, Visual Reasoning, Attention Mechanism.
\end{IEEEkeywords}

\section{Introduction}
\IEEEPARstart{T}{he} rapid advancement of deep generative models has enabled AI to produce highly realistic images, raising concerns about AI misuse, particularly in political propaganda \cite{momeni2025artificial}, financial fraud \cite{kaushik2025financial}, and social media manipulation \cite{mahony2024concerns}. Accurate and interpretable detection of AI-generated images is essential for mitigating these risks. However, traditional detection methods face two key challenges. First, AI-generated images from different models exhibit distinctive artifacts, leading to a substantial domain gap that limits the generalization ability of detectors relying on generator-specific low-level features. Second, unlike image manipulation methods that modify only specific regions, every pixel in an AI-generated image is synthesized, rendering saliency-based explainability methods \cite{liu2021two,hua2023learning,guillaro2023trufor,li2024image} ineffective. These challenges highlight the need for a generalizable and interpretable AI-generated image detector capable of understanding forgery attributes beyond model-specific patterns.
\IEEEpubidadjcol 

Prior research has attempted to address these issues using two primary strategies: 1) low-level feature analysis focuses on detecting common fingerprints present across generative models \cite{gao2023self,karageorgiou2025any,tan2024rethinking,fu2025faces}, but these methods suffer from severe domain shift, as training on one set of generative models does not generalize well to unseen model sets; 2) high-level semantics has recently been explored using Vision-Language Models (VLMs) to detect semantics inconsistencies within AI-generated images \cite{ojha2023towards,liu2024forgery,tan2025c2p}. However, due to image-text alignment pretraining, they lack sensitivity of fine-grained artifacts, leading to suboptimal detection performance. Furthermore, while MLLMs have demonstrated some ability to explain manipulated images, existing MLLM-based methods \cite{zhang2024common,huang2024ffaa,xu2024fakeshield,huang2025sida} fail to explore their integration with forgery detection task, limiting their potential in forensics analysis.

To overcome these limitations, we propose modeling AI-generated image detection and explanation as a Forgery Detection and Reasoning task (FDR-Task). As shown in Figure \ref{fig:FDR-Task}, the task leverages MLLMs to perform accurate detection through reliable reasoning over forgery attributes. To facilitate this task, we first propose the Multi-Modal Forgery Reasoning Dataset (MMFR-Dataset), a large-scale dataset containing 120K images across 10 generative models, with 378K reasoning annotations on forgery attributes. Leveraging GPT-4o and a customized construction pipeline, we generate comprehensive reasoning annotations via forgery-oriented prompts. Each annotation is organized into structured stages and hierarchical steps to form a coherent chain-of-thought. To the end, MMFR-Dataset provides a foundation for large-scale training and evaluation for MLLM-based forgery detectors.

Building upon the MMFR-Dataset, we propose FakeReasoning, a forgery detection and reasoning framework that enhances both detection accuracy and interpretability. FakeReasoning introduces three key components. First, a dual-branch visual encoder that combines CLIP and DINO is employed to extract both high-level semantics and low-level artifacts, enhancing the fine-grained forgery perception of MLLMs. Second, the Forgery-Aware Feature Fusion module (FAFF) uses DINO’s attention maps as forgery-aware priors and fuses the two-branch visual features through a cross-attention mechanism. This design adaptively guides MLLMs' attention on low-level and high-level clues while highlighting potential forgery regions and entities. Third, the Classification Probability Mapper (CPM) couples forgery detection with language modeling by mapping the vocabulary probability distribution to classification scores, effectively bridging the optimization gap and improving detection performance. Our contributions can be summarized as follows:

\begin{itemize}
    \item We formulate AI-generated image detection and explanation as a Forgery Detection and Reasoning task (FDR-Task), enabling models to jointly assess image authenticity and analyze forgery attributes. 
    \item We introduce the Multi-Modal Forgery Reasoning Dataset (MMFR-Dataset), a large-scale dataset with comprehensive reasoning annotations, facilitating both training and evaluation of MLLM-based forgery detectors.
    \item We propose FakeReasoning, a forgery detection and reasoning framework that enhances fine-grained visual perception of MLLMs and couples language modeling task with forgery detection.
    \item Extensive experiments across generative models demonstrate that FakeReasoning not only achieves robust generalization but also outperforms state-of-the-art methods in both detection accuracy and interpretability.
\end{itemize}

\section{RELATED WORK}

In this section, we review recent advances in forgery detection and multi-modal large language models, analyze the motivation for integrating MLLMs into forgery detection, and summarize the limitations of existing approaches.

\subsection{AI-Generated Image Detection} 
With the emergence of diverse generative models, it's crucial to design detectors generalizing to unseen domains. 
Traditional methods focus on uncovering low-level artifacts shared across generative models. CNNDetection \cite{wang2020cnn} demonstrates that the detector trained on ProGAN \cite{karras2017progressive} images exhibits strong generalization across GAN-generted images. 
In the meanwhile, DIRE \cite{wang2023dire} indicates that diffusion-model-generated images can be well-reconstructed through pretrained diffusion models, whereas the reconstruction of real images is hard. 
DRCT \cite{chendrct} creates hard examples by reconstructing real images and jointly conducts contrastive learning and binary classification on both original and reconstructed images. These methods achieve good intra-domain generalization but struggle to generalize across unseen domains.

Recently, high-level semantics has been explored using vision-language models to detect semantics inconsistencies within AI-generated images. UniDF \cite{ojha2023towards} uses a frozen CLIP visual encoder \cite{radford2021learning} to avoid the mode asymmetry. 
FatFormer \cite{liu2024forgery} follows zero-shot inference of CLIP, enhancing the prompt template with image embeddings and calculating cosine similarity for contrastive learning. 
Although vision-language models excel at capturing global semantic representations, it struggles to model low-level artifacts through image-text alignment pretraining.

\subsection{Multi-Modal Large Language Models} 
Multi-modal large language models have emerged as a powerful tool for vision-language understanding. 
Recent advances in MLLMs reveal several trends across representative models such as Qwen-2.5-VL \cite{Qwen2.5-VL}, GLM-4V \cite{glm2024chatglm}, and InternVL \cite{chen2024internvl}. First, there is a clear emphasis on scalable visual encoders capable of preserving high-resolution spatial information, either through dynamic resolution handling or high-capacity vision transformers. Second, these models increasingly adopt enhanced cross-modal fusion mechanisms, replacing lightweight adapters with more expressive modules, such as MLP-based mergers, gated activation functions, or large-scale query-driven cross-attention, to better align visual and textual features. Besides, DeepSeek-VL2 \cite{wu2024deepseek} integrates a Mixture-of-Experts (MoE) language model for efficient and scalable multi-modal reasoning. It employs sparsely activated expert routing, activating only a subset of parameters during inference. 

With MLLMs' strengths in cross-modal alignment and text generation, they have shown growing potential in forgery detection through detecting semantic inconsistencies and explaining suspicious visual cues.

\begin{table*}[ht]
\centering
\caption{Comparison with existing MLLM-based forgery detectors.}
\renewcommand{\arraystretch}{1.2}
\label{tab:mllm_method}
\begin{tabular}{llcccc}
\toprule
Method        & Reference    & Field              & Visual Encoder    & Reasoning Paradigm              & Detection Paradigm \\ 
\midrule
DD-VQA \cite{zhang2024common}       & 2024 ECCV    & Deepfakes          & CLIP              & Textual Explanation   & Classification Head \\
FakeShield \cite{xu2024fakeshield}   & 2025 ICLR    & Manipulation & CLIP              & Textual Explanation   & Text Generation  \\
SIDA \cite{huang2025sida}         & 2025 CVPR    & General            & CLIP              & Textual Explanation   & Classification Head \\
M2F2-Det \cite{guo2025rethinking}     & 2025 CVPR    & Deepfakes          & CLIP + Expert     & Textual Explanation   & Classification Head \\
AIGI-Holmes \cite{zhou2025aigi}  & 2025 ICCV    & General            & CLIP + Expert     & Textual Explanation   & Text Generation  \\
LEGION \cite{kang2025legion}       & 2025 ICCV    & General            & CLIP              & Textual Explanation   & Classification Head \\
\midrule
FakeReasoning & Ours         & General            & CLIP + DINO       & Chain-of-Thought      & Probability Mapping \\
\bottomrule
\end{tabular}
\end{table*}

\subsection{Explainable Forgery Detection}

To enhance the interpretability of forgery detection, localization has been introduced. Noiseprint \cite{cozzolino2019noiseprint} and TruFor \cite{guillaro2023trufor} employ noise patterns to identify tampered regions. HiFi-Net \cite{guo2023hierarchical} leverages the hierarchical relationships among generators as priors to enable fine-grained manipulation localization. However, localization-based methods often suffer from high false positive rates on authentic images. Moreover, with the rapid advancement of generative models, every pixel in an AI-generated image is synthesized, rendering saliency-based detectors ineffective.

As mentioned above, MLLMs acquire strong image understanding capabilities through large-scale image-text alignment pretraining. 
More recent MLLM-based detectors are summarized from field, visual encoder, reasoning paradigm and detection paradigm in Table \ref{tab:mllm_method}. Existing MLLM-based detectors differ in their designs of visual encoders. Several method \cite{guo2025rethinking,zhou2025aigi} introduce additional expert encoders to capture low-level artifacts, yet suffer from two limitations. First, expert encoders require an extra pretraining stage, which introduces substantial training cost. Besides, although CNN-based expert encoders excel in low-level perception, they are not aligned with the textual embedding space, thereby constraining the downstream forgery detection and reasoning task. Beyond visual representation, existing MLLM-based detectors simply use textual outputs as the explanation of detection results, failing to exploit the visual reasoning capability of MLLM and lack logical organization of forgery-related cues. From the perspective of detection paradigms, some methods \cite{zhang2024common,huang2025sida,guo2025rethinking,kang2025legion} include independent classification head to get detection results, where the classifier is decoupled from MLLMs, leading to inconsistencies between explanations and detection results. The other methods \cite{xu2024fakeshield,zhou2025aigi} simply get detection results from textual output, which ignores the optimization gap between language modeling and forgery detection, leading to suboptimal performance.

\section{Multi-Modal Forgery Reasoning Dataset}
\begin{table*}[ht]
\caption{Comparison with existing datasets. Generators denotes the number of generative models.}
\renewcommand{\arraystretch}{1.2}
\centering
\begin{tabular}{llcccccc}
\toprule
Dataset      & Reference  & Field     & Annotator & Latest Generator          & Generators & Images & Annotations \\ 
\midrule
DD-VQA \cite{zhang2024common}      & 2024 ECCV  & Deepfakes & Human     & NeuralTextures \cite{thies2019deferred} (2019) & 4      & 3K     & 14K         \\
FFA-VQA \cite{huang2024ffaa}      & 2024 arXiv & Deepfakes & GPT-4o    & GANDiffFace \cite{melzi2023gandiffface} (2023)    & 7      & 95K    & 20K         \\
Fakebench \cite{li2024fakebench}    & 2024 arXiv & General   & GPT-4V    & DALLE3 \cite{betker2023improving} (2023)         & 10     & 3K     & 57K         \\
LOKI \cite{ye2024loki}         & 2025 ICLR  & General   & Human     & FLUX \cite{flux2024} (2024)           & 10     & 2K     & 0.2K         \\
MMTD-Set \cite{xu2024fakeshield}     & 2025 ICLR  & Manipulation   & GPT-4o    & RePaint \cite{lugmayr2022repaint} (2022)        & 2      & 150K   & 34K         \\
SID-Set \cite{huang2025sida}      & 2025 CVPR  & General   & GPT-4o    & FLUX \cite{flux2024} (2024)           & 2      & 300K   & 3K          \\
SynthScars \cite{kang2025legion}   & 2025 arXiv & General   & Human     & DALLE3 \cite{betker2023improving} (2023)         & 10     & 12K    & 26K         \\
\midrule
MMFR-Dataset & Ours       & General   & GPT-4o    & GPT-4o (2025)         & 10     & 120K   & 378K        \\ 
\bottomrule
\end{tabular}
\label{tab:dataset comparison}
\end{table*}

\begin{figure*}
    \centering
    \includegraphics[width=\textwidth]{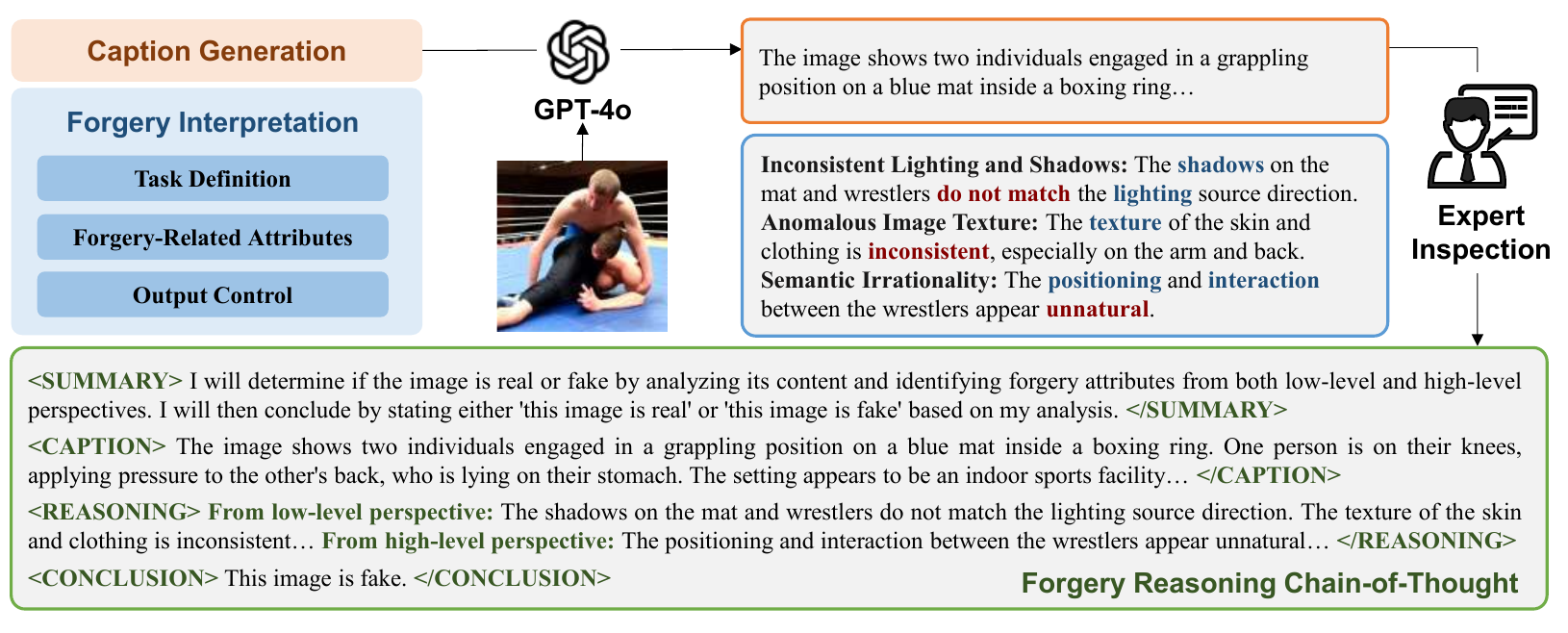}
    \caption{Construction pipeline of the MMFR-Dataset. GPT-4o is tasked with caption generation and forgery interpretation. For the forgery interpretation task, the prompt is crafted to instruct GPT-4o to analysis forgery-related attributes. After inspected by human experts, the generated captions and interpretations are compiled with a chain-of-thought to enhance structured and hierarchical reasoning.}
    \label{fig:dataset_pipeline}
\end{figure*}

In the vision–language community, recent reasoning datasets focus on language-guided visual understanding. For instance, MRSeg \cite{wang2024segllm} introduces multi-round reasoning for segmentation, while gRefCOCO \cite{liu2023gres} and MeViS \cite{ding2023mevis} extend referring expression segmentation to more generalized and motion-aware scenarios. These benchmarks emphasize compositional and contextual grounding in generic settings.
To perform the proposed FDR-Task, high-quality images with forgery-related annotations are essential. Recently, several datasets have been developed to evaluate \cite{ye2024loki,li2024fakebench} or fine-tune MLLMs \cite{zhang2024common,huang2024ffaa,xu2024fakeshield,huang2025sida,kang2025legion}.
However, as summarized in Table \ref{tab:dataset comparison}, existing datasets still face several challenges: 1) Insufficient annotations: current datasets either serve as benchmarks which lack enough samples for training or contain far fewer annotations compared to images, limiting the exploitation of MLLMs' forensics capabilities. 2) Object-centric focus: datasets such as MMTD-Set \cite{xu2024fakeshield} and parts of SID-Set \cite{huang2025sida} target tampered images with explicit manipulated masks, which neglecting holistic image analysis and limiting model generalizability.

To address these challenges, we propose the MMFR-Dataset, a large-scale synthetic image dataset with detailed annotations over forgery attributes. Leveraging GPT-4o and a customized data construction pipeline, we generate a substantial number of samples with comprehensive reasoning annotations. Each annotation is further organized in a structured (four-stage) and hierarchical (low-level and high-level) format, enabling a forgery reasoning chain-of-thought.

\subsection{Source Image Collection}

Training set:  We follow the widely used ``train-on-one and test-on-many'' paradigm \cite{yang2025d} in AI-generated image detection \cite{wang2020cnn,ojha2023towards}. For fake images, we choose DiffusionDB dataset \cite{wang2022diffusiondb}. Different from large-scale, automatically generated datasets, DiffusionDB collects well-crafted images generated by Stable Diffusion users on Discord, ensuring both generation quality and content diversity. For real images, we use LAION-Aesthetics v2 dataset \cite{schuhmann2022laion}. 

Evaluation sets: In order to comprehensively evaluate the generalization of forgery detectors, we curate a collection of recent open-source AI-generated image datasets. 
Each evaluation set comprises 1,000 fake images generated by state-of-the-art models released within the past three years, including: Stable Diffusion, DALLE-3, DeepFloyd IF, Midjourney, Kandinsky, PixArt-$\alpha$, FLUX, GPT-4o, StyleGAN-XL, and GigaGAN. 
To ensure distributional consistency between real and fake images, we also collect 1,000 real images from ImageNet \cite{imagenet15russakovsky} and 1,000 from LAION \cite{schuhmann2022laion}. ImageNet serves as the real set for StyleGAN-XL, while LAION is used for the others. 

\subsection{Forgery Reasoning Generation}

Forgery reasoning annotations in the MMFR-Dataset is generated in two steps. First, we design a \textit{Forgery Interpretation Prompt} and employ GPT-4o to interpret specific attributes signifying the image authenticity. Second, we decompose the process into four structured stages, with the reasoning stage further divided into two perspectives. The construction flow is shown in Figure \ref{fig:dataset_pipeline}.

\subsubsection{GPT-4o Assisted Forgery Interpretation}
GPT-4o has proven its capabilities in detecting and interpreting potential forgeries within images \cite{zhang2024bench,li2024fakebench}. Compared with manual analysis, employing GPT-4o for automated generation significantly expands the size of datasets, which is widely adopted in instruction tuning \cite{liu2024visual} and domain-specific fine-tuning \cite{xu2024fakeshield,tran2024bioinstruct}. To generate precise and comprehensive analysis of forgery-related attributes, we design a \textit{Forgery Interpretation Prompt} considering following aspects:

\noindent \textbf{Task definition.} First of all, the \textit{Prompt} specifies the image authenticity as either ``real'' or ``fake''. GPT-4o is tasked with identifying specific attributes that signify the image authenticity and providing detailed descriptions.

\noindent \textbf{Forgery-related attributes.} The \textit{Prompt} includes a comprehensive set of attributes widely used in image forensics, ranging from low-level artifacts to high-level semantics, instructing GPT-4o to focus on specific features indicative of image authenticity.

\noindent \textbf{Output Control.} The \textit{Prompt} defines a json format including not only forgery attributes but also the authenticity label, which serves to enhance GPT-4o’s analysis on specific types. Besides, the \textit{Prompt} also specifies a distinct output for failure cases where GPT-4o is unable to ascertain the image authenticity. Such outputs are filtered out after generation. 
As a result, approximately 0.7K images are filtered due to label inconsistency, inability to determine authenticity, or invalid JSON structure during automatic filtering, preventing potential bias in forgery detection task.

\begin{figure*}[t]
\begin{center}
  \includegraphics[width=0.95\linewidth]{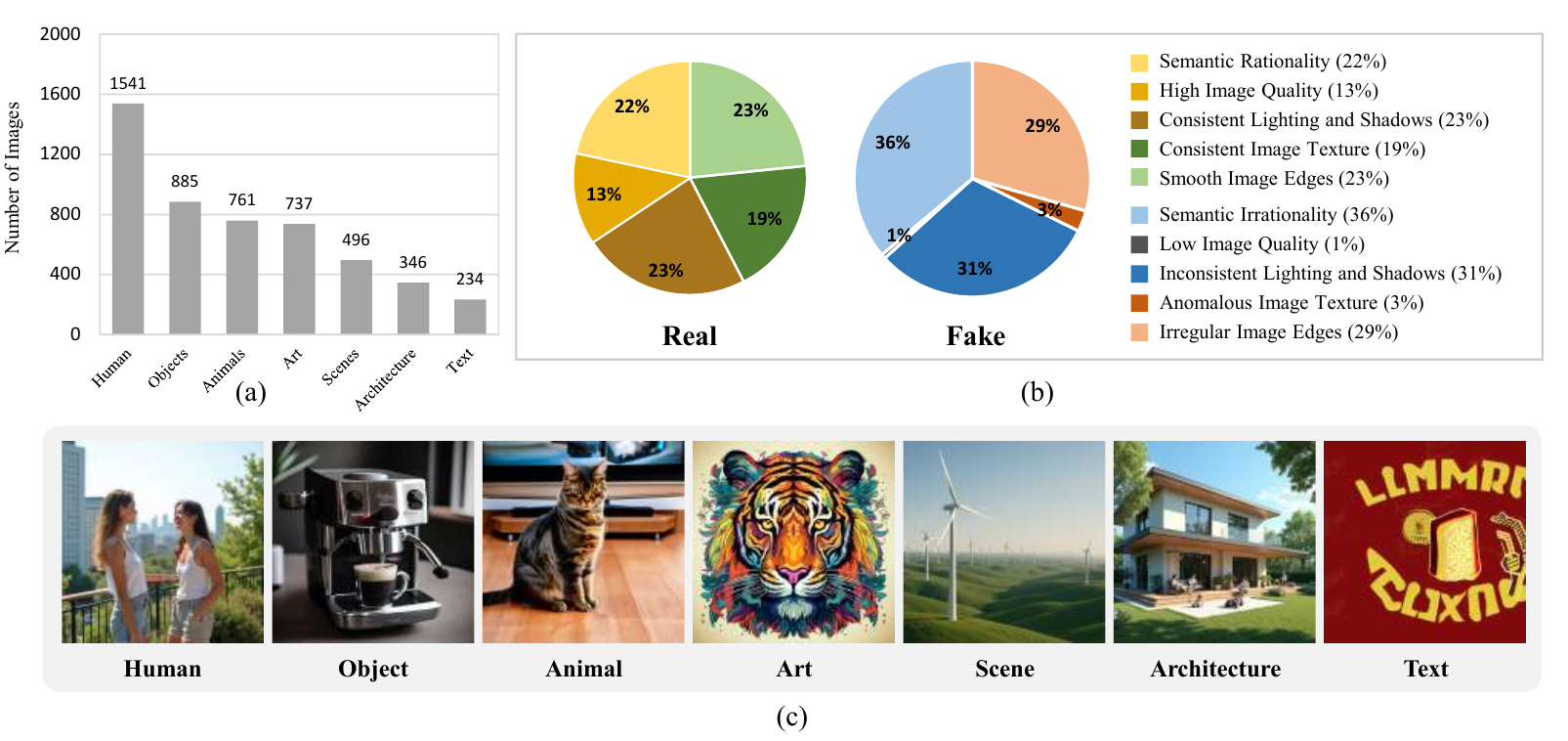}
\end{center}
  \caption{Statistics of the MMFR-Dataset. (a) Class distribution of the evaluation sets. (b) Attributes distribution of real and fake images. (c) Representative samples of the evaluation sets.}
  \label{fig:dataset_statistics}
\end{figure*}

\subsubsection{Expert Verification}

To ensure annotation reliability and reduce hallucinations, we further introduce an expert verification stage. Specifically, experts evaluate the given reasoning annotations from two dimensions, accuracy and visibility.  Each annotation is screened by a expert answering a single-choice question: ``Does this annotation accurately describe the forgery-related attributes visible in the image?'' with three options:  ACCEPT / REJECT / UNSURE. All annotations are independently reviewed in the first round. Annotations labeled as UNSURE are re-evaluated in a second round by three experts, with final outcomes determined through majority voting between ACCEPT / REJECT decisions. 

The expert team consists of 10 researchers specializing in image forensics, including graduate students, Ph.D. students, and researcher scientists. On average, each expert requires approximately 15 seconds to verify one image with its reasoning annotations. And the complete process consumes around 530 human-hours per expert. After two rounds of verification, the average number of reasoning annotations per image decreases from 3.75 to 3.15, corresponding to a rejection rate of 16\%. And approximately 1,000 images are removed because because all of their annotations are rejected.

Since the first-round verification is conducted by a single expert, we perform an additional blind spot check to assess reliability. Specifically, we first select the 500 fastest and 500 slowest images in the first round. After excluding these samples, we randomly select 1,000 images. In total, 2,000 images with 7.8K reasoning annotations are evaluated by an image forensics expert who is not involved in dataset construction and follows the same criteria. Among all the annotations, the agreement rate reaches 94.87\%, with false acceptance and false rejection rates of 4.62\% and 7.69\%, respectively. These results indicate strong consistency and effective filtering of hallucination annotations.

\subsubsection{Forgery Reasoning Chain-of-Thought}
Inspired by recent advances in multi-modal chain-of-thought \cite{shao2024visual,xu2024llava,diao2025driverx}, we develop a specific chain-of-thought for forgery reasoning task, consisting of structured stages and hierarchical steps, to enhance forgery reasoning capability of MLLMs. First, we decompose the whole process into four stages:

\noindent \textbf{Summary.} In this stage, the \textit{Forgery Reasoning CoT} provides a brief summary of the FDR-Task and explains the steps to conduct the whole process.

\noindent \textbf{Caption.} The \textit{Forgery Reasoning CoT} gives a description of elements and relations within an image to help understand image and find forgeries.

\noindent \textbf{Reasoning.} The interpretations generated by GPT-4o are further organized into low-level and high-level reasoning perspectives based on the types of forgery attributes. Note that, since current MLLMs struggle with low-level perception, the hierarchical reasoning steps in the \textit{Forgery Reasoning CoT} provide an effective supervision to the feature extraction of the dual-branch visual encoder in the proposed method.

\noindent \textbf{Conclusion.} Based on above stages, the \textit{Forgery Reasoning CoT} outputs the final decision on image authenticity.

 To emphasize structured reasoning stages, we use four pairs of special \texttt{<SEG>} tags: {\small \texttt{<SUMMARY></SUMMARY>}}, {\small \texttt{<CAPTION></CAPTION>}}, {\small \texttt{<REASONING></REASONING>}} and {\small \texttt{<CONCLUSION></CONCLUSION>}} at both beginning and end of each stage. 

 \subsection{Dataset Statistics}
 
 The training set of the MMFR-Dataset consists of 50,266 synthetic images with 129,345 reasoning annotations, and 50,990 real images with 183,082 reasoning annotations. Across all annotations, there are 10 types of forgery-related attributes, whose distributions in real and fake images are depicted in Figure \ref{fig:dataset_statistics}.

 The evaluation sets of the MMFR-Dataset comprise 10,000 synthetic images with 25,054 reasoning annotations and 10,000 real images with 41,726 reasoning annotations, providing foundation for comprehensive evaluation of the FDR-Task. To characterize the semantic composition of the evaluation sets, we analyze the prompts used to generate evaluation sets which are publicly available, including DiffusionDB \cite{wang2022diffusiondb}, Kandinsky \cite{diffusers_parti_prompts_kandinsky2_2_2023}, PixArt \cite{pixart_alpha_pixart_eval30k_2024}, FLUX \cite{lehduong_flux_generated_2025}, and GPT-4o \cite{chen2025_sharegpt4o_image}. A total of 5,000 prompts are categorized into high-level semantic classes using GPT-4o, namely Human, Objects, Animals, Art, Scenes, Architecture, and Text. The class distribution and representative samples are shown in Figure \ref{fig:dataset_statistics}.

\section{FakeReasoning}

\begin{figure*}
    \centering
    \includegraphics[width=\textwidth]{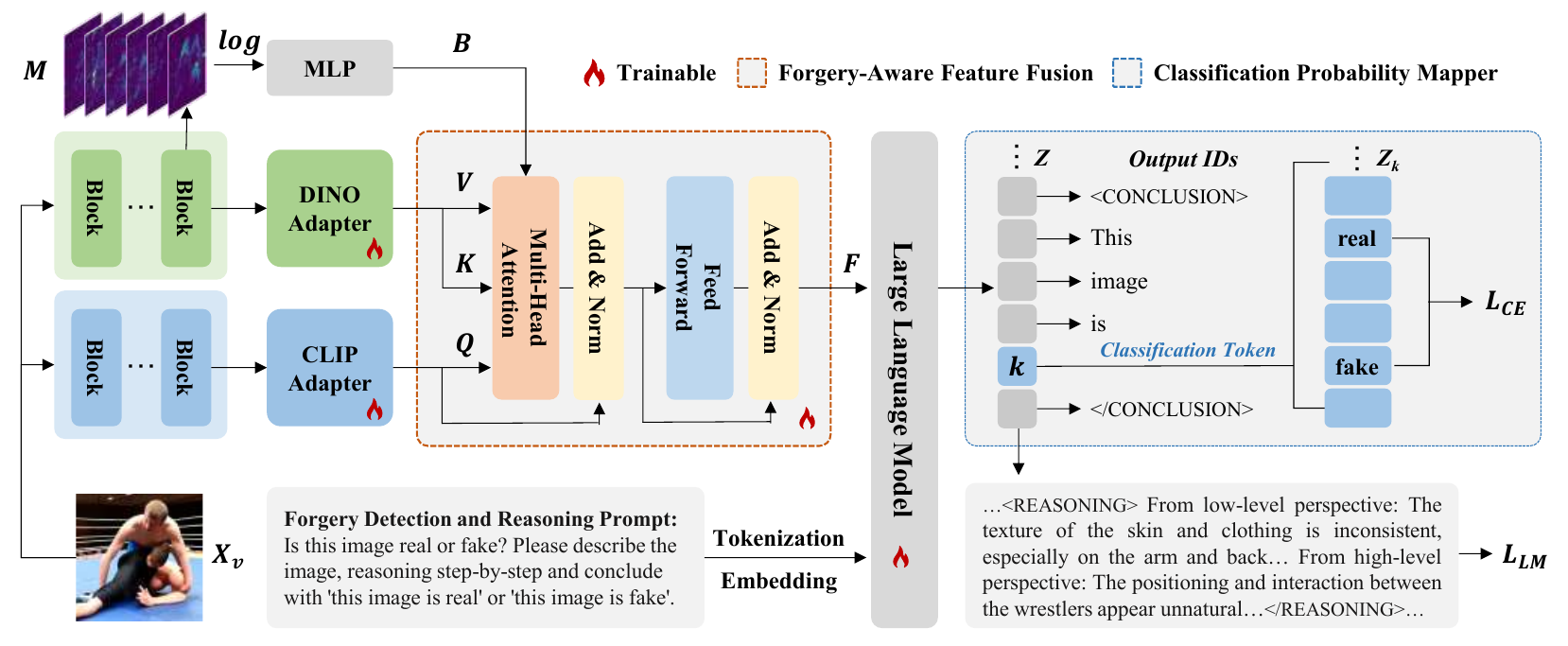}
    \caption{The pipeline of FakeReasoning. FakeReasoning adopts a dual-branch visual encoder combining CLIP and DINO to extract both high-level and low-level visual clues. Each encoder is followed by an adapter to align with the text embedding. The Forgery-Aware Feature Fusion module further fuses CLIP tokens and DINO tokens with cross-attention mechanism, leveraging DINO's attention maps as forgery-aware priors. The Classification Probability Mapper locates the classification token indicating the image authenticity from original logits and maps the vocabulary probability distribution into a classification score.}
    \label{fig:fakereasoning}
\end{figure*}

The pipeline of FakeReasoning is illustrated in Figure \ref{fig:fakereasoning}. To enhance the fine-grained perception of MLLMs, FakeReasoning adopts a dual-branch visual encoder combining CLIP and DINO to extract both high-level and low-level visual clues. Each encoder is followed by an adapter to align with the text embedding. The Forgery-Aware Feature Fusion module (FAFF) further fuses CLIP tokens and DINO tokens with cross-attention mechanism, leveraging DINO's attention maps as forgery-aware priors. This design adaptively guides MLLMs’ attention on low-level and high-level clues while highlighting potential forgery regions and entities. The Classification Probability Mapper (CPM) locates the classification token indicating the image authenticity from original logits and maps the vocabulary probability distribution into a classification score, which effectively couples language modeling with forgery detection and improves overall performance.

\subsection{Dual-Branch Visual Encoder}

To analyze the authenticity of an image, both low-level features \cite{zhong2023patchcraft,dong2022mvss,lai2024lideepdet} and high-level semantics are essential \cite{tan2025c2p,ojha2023towards,liu2024forgery}. Most existing open-source MLLMs \cite{liu2024visual, dai2023instructblipgeneralpurposevisionlanguagemodels,chen2023minigptv2} adopt pretrained CLIP as their visual encoder. Although CLIP encoder excels at capturing global semantic representations, it lacks sensitivity to low-level visual clues due to image-text alignment pretraining \cite{luo2023controlling,cai2025computer,he2024rigid}, which poses a major limitation for MLLMs in image forensics.

Meanwhile, DINO \cite{caron2021emerging} is pretrained through vision-only self-supervised learning, demonstrating strong capability in local perspective. Prior work has shown that integrating DINO features with CLIP features enhances fine-grained perception and mitigates hallucinations of MLLMs \cite{tong2024eyes,jiang2023clip}. Inspired by these advances, we further explore this ``feature complementarity effect'' in MLLM-based image forensics.

Specifically, the image \(X_v\) is input into a frozen CLIP and DINO visual encoder to obtain visual embeddings. Since the the CLIP encoder is pretrained with text supervision while the DINO encoder is not, we introduce separate MLP modules, $A_C$ for CLIP and $A_D$ for DINO, as adapters to align with textual embedding space. These two adapters account for the heterogeneous nature of the two encoders and facilitate more effective modality alignment. Accordingly, the visual tokens from CLIP $F_C$ and DINO $F_D$ are obtained as follows:
\begin{align}
F_C &= A_C\big(\operatorname{CLIP}(X_v)\big),\\
F_D &= A_D\big(\operatorname{DINO}(X_v)\big).
\end{align}

\subsection{Forgery-Aware Feature Fusion}
Fusion strategies of MLLMs' visual embeddings can be broadly categorized into three types: feature weighting, feature concatenation, and layer interaction.
Feature weighting aims to assign importance scores to different visual embeddings, yet it suffers from the distributional mismatch between CLIP and DINO representations, and can degrade the instruction-following ability of pretrained MLLMs \cite{tong2024eyes}.
Feature concatenation directly merges different visual embeddings into a single representation, lacking the adaptability to guide MLLMs' attention on complementary low-level artifacts and high-level semantics.
Layer interaction methods facilitate cross-layer communication between the two visual branches, but they typically introduce substantial computational complexity and rely on manually selected layers \cite{jiang2023clip,luo2023controlling}. 

To address these limitations and obtain discriminative forgery features, we introduce a Forgery-Aware Feature Fusion module, which leverages DINO's attention maps as forgery-aware priors and applies a cross-attention mechanism to integrate CLIP and DINO visual representations. 
To start with, the final-layer attention maps $M \in \mathbb{R}^{n \times H \times L \times L}$ of the DINO visual encoder highlights $H$ regions rich in semantic and entity-level information \cite{oquab2023dinov2}. 
Leveraging this property, we use the attention maps $M$ as forgery-aware priors and incorporate them into the cross-attention mechanism as the attention bias, highlighting potential forgery regions and entities. The attention bias $B$ is computed as:
\begin{equation}
B = \operatorname{MLP}(\log M),
\end{equation}

\noindent where the logarithmic transformation aligns the scale of softmax-normalized attention maps and similarity logits \cite{vaswani2017attention}. The subsequent lightweight $\operatorname{MLP}$ contains two layers with a GELU non-linearity, which serves as a non-linear projection module that increases the representational capacity and enables head-specific modulation \cite{dosovitskiy2020image}. 

Using $B$ as the attention bias, we take CLIP tokens $F_C \in \mathbb{R}^{n \times L \times d}$ as queries and DINO tokens $F_D \in \mathbb{R}^{n \times L \times d}$ as keys and values for cross-attention. 
Incorporating forgery-aware priors into fused feature $F$, the cross-attention is defined as:
\begin{align}
A &= \operatorname{softmax}\!\Big(\frac{F_C F_D^\top}{\sqrt{d}} + B\Big),\\
F' &= \operatorname{LN}\big(A F_D + F_C\big),\\
F &= \operatorname{LN}\big(\operatorname{FFN}(F') + F'\big),
\end{align}

\noindent where $F_D^\top$ represents the transpose of the DINO tokens and $\operatorname{FFN}$ denotes the feed-forward network. Notably, the FAFF module employs $H$ heads for cross-attention, consistent with the architecture of the CLIP and DINO visual encoders.

\subsection{Classification Probability Mapper}
In this section, we aim to explore the integration of MLLMs and forgery detection task. Existing MLLM-based detectors can be categorized into two types: 1) optimizing original language modeling task to indirectly improve detection accuracy \cite{keita2025bi,chang2023antifakeprompt,xu2024fakeshield,zhang2024common}; 2) training an independent classification head with intermediate features from MLLMs \cite{huang2024ffaa,guo2025rethinking,huang2025sida}. The first category overlooks the objective discrepancy between language modeling and forgery detection, resulting in suboptimal detection performance. And the second category, on the other hand, ignores the intrinsic connections between these two tasks, which can cause inconsistencies between the textual interpretation and the classification result, thereby undermining the detector's credibility.

To leverage the intrinsic connections between language modeling and forgery detection, we propose the Classification Probability Mapper. As illustrated in Figure \ref{fig:fakereasoning}, CPM locates the classification token indicating the image authenticity from original logits and maps the vocabulary probability distribution into a classification score. Specifically, the original logits $Z \in \mathbb{R}^{n \times N \times V}$ output from MLLMs represent the predicted probabilities of each token in the vocabulary, where \( N \) is the length of the output sequence and \( V \) is the size of the vocabulary. Using the Greedy Search Strategy \cite{sutskever2014sequence}, we sample the predicted words based on the highest probabilities and obtain the output IDs:
\begin{equation}
O = \arg\max_{v} (Z).
\end{equation}

\noindent where the $\arg\max$ is taken over the vocabulary dimension $v \in \{0, 1, \ldots, V-1\}$ to obtain the output IDs $O \in \mathbb{R}^{n \times N}$.

According to the \textit{Forgery Reasoning CoT}, image authenticity is determined in the conclusion stage. Therefore, we encode \texttt{<CONCLUSION>} and answer template ``This image is'' into token IDs as the match pattern \( p \). We use this pattern to search through the output IDs and identify the index \( k \) of classification token:
\begin{equation}
k =  \min\{ i + l \mid O_{i:i+l} = p, \, i \in [0, N - l] \},
\end{equation}

\noindent where $l$ is the length of the pattern.

For the classification token, its logits corresponding to the vocabulary entries ``real'' and ``fake'' determine the image authenticity. Therefore, we model the forgery detection as a special case of multi-class classification, which is optimized with the cross-entropy loss:
\begin{equation}
\mathcal{L}_{CE}=-\log\frac{\exp\left(Z_{k,ID_{y}}\right)}{\exp\left(Z_{k,ID_{\mathrm{real}}}\right)+\exp\left(Z_{k,ID_{\mathrm{fake}}}\right)},
\end{equation}

\noindent where $y \in \{ \mathrm{real}, \mathrm{fake} \}$ denotes the ground-truth label of image authenticity, $Z_{k,ID_{\mathrm{real}}}$ and $Z_{k,ID_{\mathrm{fake}}}$ denote the original logits of the classification token projected onto the vocabulary entries ``real'' and ``fake''. 
To balance the scale of losses, we introduce a temperature hyperparameter $\tau$ to original logits in $\mathcal{L}_{CE}$ by $Z / \tau$, which is set to 10 in our experiment.

Besides, to enhance the robustness of pattern matching and optimization, a masking mechanism is applied during the $\mathcal{L}_{CE}$ computation. Specifically, the index of unmatched samples is set to $k = -1$, and only the losses of matched samples are included in backpropagation. When all samples within a batch fail to match, the corresponding $\mathcal{L}_{CE}$ drops to zero. Furthermore, the multi-GPU parallelism and gradient accumulation are used to expand the effective batch size, leading to more stable optimization.

\subsection{Joint Optimization}

We use images and forgery reasoning annotations from the MMFR-Dataset as the visual input \(X_v\) and answers \(X_a\). The question \(X_q\), such as the Forgery Detection and Reasoning Prompt, is defined in Figure \ref{fig:fakereasoning}. Following the methodology of LLaVA \cite{liu2024visual}, the language modeling task is optimized using an auto-regressive training objective:
\begin{equation}
\mathcal{L}_{LM} = -\sum_{j=1}^N \log p_\theta(x_j|X_v, X_{q,<j}, X_{a,<j}),
\end{equation}

\noindent where \(\theta\) represents the trainable parameters, \(X_{\mathrm{q},<j}\) and \(X_{\mathrm{a},<j}\) denote the question and answer tokens from all previous turns prior to the current prediction token \(x_j\), and \(N\) is the sequence length. During training, FakeReasoning is optimized through joint losses:
\begin{equation}
\text{Loss} = \mathcal{L}_{CE} + \mathcal{L}_{LM}.
\end{equation}

\section{Experiment}

\begin{table*}[t]
\caption{Comparison on forgery detection task. ${\dagger}$ denotes using official weights and the others are trained on the MMFR-Dataset. \\SD, IF, Mj, Kd, and SG-XL represent Stable Diffusion, DeepFloyd IF, Midjourney, Kandinsky and StyleGAN-XL.}
\centering
\renewcommand{\arraystretch}{1.2} 
\begin{tabular}{
  l@{\hskip 10pt}  
  l@{\hskip 13pt}   
  *{10}{c@{\hskip 10pt}}  
  c                
}
\toprule
Method & Reference & SD & DALLE-3 & IF & Mj & Kd & PixArt & FLUX & GPT-4o & SG-XL & GigaGAN & AVG \\ 
\midrule
UniFD \cite{ojha2023towards} & 2023 CVPR & 92.90 & \underline{88.80} & 82.90 & 80.80 & 81.90 & 88.70 & 81.00 & 91.80 & 72.70 & 86.50 & 84.80 \\
DE-FAKE \cite{sha2023fake} & 2023 CCS & 96.25 & 80.00 & 82.50 & 63.75 & 86.25 & 90.00 & \underline{86.25} & 95.00 & 61.25 & 57.50 & 79.88 \\
FatFormer$^{\dagger}$ \cite{liu2024forgery} & 2024 CVPR & 79.65 & 52.40 & 63.70 & 55.24 & 50.91 & 65.05 & 37.56 & 50.04 & \textbf{97.40} & 84.70 & 63.67 \\
NPR \cite{tan2024rethinking} & 2024 CVPR & 98.80 & 50.90 & 72.60 & \underline{94.60} & 91.90 & 86.50 & 49.90 & 74.40 & 79.80 & 77.90 & 77.73 \\
DRCT$^{\dagger}$  \cite{chendrct} & 2024 ICML & 75.30 & 70.40 & 71.90 & 75.76 & 65.31 & 75.56 & 81.71 & 74.81 & 82.65 & 58.65 & 73.21 \\
FreqNet \cite{tan2024frequency} & 2024 AAAI & \textbf{99.30} & 49.80 & 53.30 & 93.90 & 77.20 & 91.00 & 49.90 & 82.30 & 95.10 & 80.30 & 77.21 \\
C2P$^{\dagger}$ \cite{tan2025c2p} & 2025 AAAI & 83.30 & 61.70 & 54.75 & 56.71 & 58.93 & 50.24 & 39.61 & 42.42 & \underline{96.70} & 85.50 & 62.99 \\
AIDE \cite{yan2024sanity} & 2025 ICLR & 96.75 & 88.55 & 73.65 & 68.19 & 72.53 & 83.26 & 86.08 & 95.51 & 84.15 & 77.10 & 82.58 \\
SPAI \cite{karageorgiou2025any} & 2025 CVPR & 98.10 & 76.85 & 75.85 & 68.20 & 65.20 & 68.30 & 58.25 & 66.75 & 66.15 & 56.95 & 70.06 \\
Effort \cite{yan2024orthogonal} & 2025 ICML & 97.90 & 79.60 & 84.40 & 80.50 & 60.70 & 93.20 & 
75.00 & 94.50 & 94.20 & 79.60 & 83.96 \\
LOTA \cite{wang2025lota} & 2025 ICCV & 98.20 & 50.10 & 95.05 & 98.97 & 98.79 & 95.49 & 50.05 & 97.45 & 93.55 & 98.25 & 87.59 \\
\midrule
FakeReasoning & Ours & \underline{99.20} & \textbf{91.28} & \textbf{99.10} & 87.68 & \textbf{96.90} & \textbf{93.49} & \textbf{87.16} & \textbf{97.80} & 90.00 & \textbf{98.00} & \textbf{94.06} \\
\bottomrule
\end{tabular}
\label{tab:forgery detection}
\end{table*}

\begin{table}[t]
\centering
\caption{Comparison on the FDR-Task. We employ GPT-4o-assisted forgery interpretations as ground truth and assess both detection and reasoning tasks.}
\renewcommand{\arraystretch}{1.2}
\begin{tabular}{l@{\hskip 6pt}c@{\hskip 6pt}c@{\hskip 6pt}c@{\hskip 6pt}c@{\hskip 6pt}c@{\hskip 6pt}c@{\hskip 6pt}c@{\hskip 6pt}}
\toprule
\multirow{2}{*}{Method} & \multicolumn{4}{c}{Detection} & \multicolumn{3}{c}{Reasoning} \\
\cmidrule(lr){2-5} \cmidrule(lr){6-8}
 & DM & GAN & AVG & Fail $\downarrow$ & BL-1 & R-L & CSS \\
\midrule
LLaVA-1.5-13B \cite{liu2023improvedllava} & \underline{73.38} & 70.85 & \underline{72.87} & \textbf{0.00} & 0.23 & 0.20 & 0.54 \\
Qwen-2.5-VL-7B \cite{Qwen2.5-VL} & 63.58 & 54.37 & 61.74 & 3.70 & 0.22 & 0.19 & 0.56 \\
InternVL-2.5-8B \cite{chen2024internvl} & 67.72 & 60.02 & 66.18 & 6.20 & 0.14 & 0.16 & 0.58 \\
GLM-4V-9B \cite{glm2024chatglm} & 68.91 & 67.31 & 68.59 & 3.35 & 0.11 & 0.14 & 0.61 \\
Deepseek-VL2 \cite{wu2024deepseek} & 68.46 & 67.59 & 68.29 & \underline{0.70} & \underline{0.24} & \underline{0.21} & \underline{0.64} \\
\midrule
FakeShield \cite{xu2024fakeshield} & 52.00 & \underline{80.00} & 57.60 & 1.30 & 0.12 & 0.16 & 0.45 \\
SIDA \cite{huang2025sida} & 51.64 & 52.84 & 51.88 & 7.00 & 0.15 & 0.16 & 0.53 \\
LEGION \cite{kang2025legion} & 50.00 & 50.00 & 50.00 & 1.20 & 0.16 & 0.18 & 0.59 \\
\midrule
FakeReasoning & \textbf{94.08} & \textbf{94.00} & \textbf{94.06} & \textbf{0.00} & \textbf{0.37} & \textbf{0.29} & \textbf{0.76} \\
\bottomrule
\end{tabular}
\label{tab:forgery detection and reasoning}
\end{table}

In this section, we present a series of experiments to validate the effectiveness of the proposed method. We begin with implementation details, followed by the introduction of state-of-the-art baselines and evaluation metrics. Subsequently, we compare and analyze the results on two benchmark datasets. Finally, we provide visualization results and ablation studies to further demonstrate the strengths and limitations of our method.

\subsection{Implementation Details}  
We adopt LLaVA-1.5-13B \cite{liu2023improvedllava} as the base MLLM, and employ ViT-L/14 variants of CLIP \cite{radford2021learning} and DINO \cite{caron2021emerging} as visual encoders. Our model is initialized with pretrained weights of LLaVA-MoF \cite{tong2024eyes} for the LLM and the two adapters. During training stage, the visual encoders are frozen, while the LLM is fine-tuned using LoRA \cite{hu2021lora} with a rank of 128 and a scaling factor of 256. The adapters and the FAFF module are fully trainable. Our model is trained for 3 epochs on 8× NVIDIA A800 40GB GPUs with a batch size of 8. During inference, the temperature is set to 0.2.

\subsection{Baseline models}

We compare the detection performance of FakeReasoning and traditional detectors from three categories: 1) CLIP-based methods, including UniFD \cite{ojha2023towards}, DE-FAKE \cite{sha2023fake}, Fatformer \cite{liu2024forgery}, C2P \cite{tan2025c2p}, and Effort \cite{yan2024orthogonal}; 2) methods based on low-level features, such as NPR \cite{tan2024rethinking}, DRCT \cite{chendrct}, FreqNet \cite{tan2024frequency}, and SPAI \cite{karageorgiou2025any}; and 3) hierarchical method AIDE \cite{yan2024sanity} that integrates high-level and low-level features. For all baseline models, we either retrain them on the MMFR-Dataset or use their officially released weights.

Also, we compare detection and reasoning performance among FakeReasoning, open-source MLLMs and MLLM-based detectors. Specifically, open-source MLLMs include LLaVA-1.5-13B \cite{liu2023improvedllava}, Qwen-2.5-VL-7B \cite{Qwen2.5-VL}, InternVL-2.5-8B \cite{chen2024internvl}, GLM-4V-9B \cite{glm2024chatglm}, Deepseek-VL2 \cite{wu2024deepseek}, while the MLLM-based detector FakeShield \cite{xu2024fakeshield} is also included for comparison. All baseline models are reproduced and evaluated using their official weights.

\subsection{Evaluation Metrics} 
For the forgery detection task, we report image-level accuracy (ACC), which is calculated based on the authenticity decisions in the conclusion stage. 
For the forgery reasoning task, we use GPT-4o-generated interpretations as the ground truth and evaluate performance using BLEU-1 \cite{li2024fakebench}, ROUGE-L \cite{li2024fakebench}, and Semantic Similarity (CSS) \cite{xu2024fakeshield} metrics. The reasoning stage output of FakeReasoning is used for comparison.

\subsection{Benchmark for FDR-Task}

\subsubsection{Comparison on Forgery Detection Task} We first conduct a comprehensive comparison on the generalization of detection task. As shown in Table \ref{tab:forgery detection}, the results include image-level accuracy for each evaluation set.

Compared to low-level-based methods such as NPR , DRCT, FreqNet, and SPAI, CLIP-based methods like UniFD, AIDE, DE-FAKE and Effort exhibit stronger cross-domain generalization. Despite being trained only on Stable Diffusion images, these models effectively detect GAN-generated contents. 
Low-level clues, e.g., frequency artifacts in FreqNet and SPAI or upsampling traces in NPR, vary across different generators, while the reconstruction error in DRCT is specific for diffusion models. 
In contrast, CLIP leverages strong semantic priors to detect high-level inconsistencies, resulting in robust generalization.
Trained on ProGAN images, FatFormer and C2P perform well on GAN-generated evaluation sets but suffer a substantial drop faced with diffusion models. 
The most recent LOTA \cite{wang2025lota} performs consistently well on some of the generative models, but exhibits a significant performance drop on DALLE-3 and FLUX. These two models involve complex post-processing and reconstruction stages, which attenuate the low-bit-plane noise patterns that LOTA \cite{wang2025lota} relies on. Consequently, LOTA \cite{wang2025lota} fails to select discriminative noise regions and degenerates to random predictions.

Besides, distinguishing fake images of FLUX and Midjourney proves to be more difficult. These commercial models have achieved a high degree of photorealism, posing increasing challenges to existing detection methods. We notice that when trained on Stable Diffusion images, several methods generalize surprisingly well on GPT-4o-generated images. We attribute this trend to similar distributions of two generators.

Meanwhile, our method outperforms all state-of-the-art baselines, with an average accuracy exceeding the suboptimal method LOTA \cite{wang2025lota} by 6.97\%. Specifically, our method achieves the highest accuracy in 7 out of 10 generative models and shows the smallest variation. With powerful prior knowledge and task-specific architecture, our method learns forgery semantics from both low-level and high-level visual clues, instead of generator-specific artifacts, which contributes to outstanding accuracy and strong generalization of our method.

\begin{figure*}[t]
    \centering
    \includegraphics[width=\textwidth]{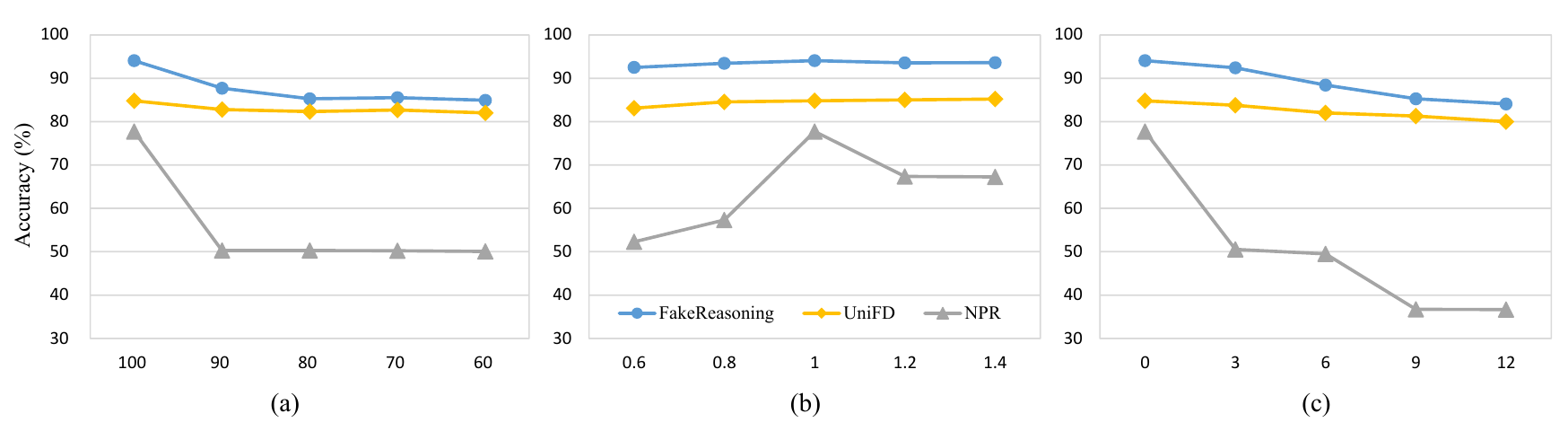}
    \caption{Robustness experiment under real-world image degradations. (a) Robustness over JPEG compression. (b) Robustness over resizing. (c) Robustness over Gaussian noise.}
    \label{fig:robust_exp}
\end{figure*}

\subsubsection{Comparison on FDR-Task} 
To establish a foundational study for the FDR-Task, we employ GPT-4o-assisted forgery interpretations as the ground truth and assess the quality of forgery reasoning. We utilize the Forgery Detection and Reasoning Prompt, as illustrated in Figure \ref{fig:fakereasoning}, to query open-source MLLMs and MLLM-based detector. The results are presented in Table \ref{tab:forgery detection and reasoning}. The ``Fail'' denotes cases where the MLLMs are unable to determine the authenticity of an image or fail to output in the required format. 

Among open-source MLLMs, LLaVA-1.5-13B \cite{liu2023improvedllava} demonstrates strong forgery detection capabilities and robust instruction-following behavior, while DeepSeek-VL2 \cite{wu2024deepseek} excels in forgery reasoning task. However, the overall performance of open-source MLLMs remains unsatisfactory. They suffer from domain shifts and uneven capabilities, limiting their applicability in real-world scenarios.

We then compare our method with recent MLLM-based detectors, including FakeShield~\cite{xu2024fakeshield}, SIDA~\cite{huang2025sida}, and LEGION~\cite{kang2025legion}. For FakeShield~\cite{xu2024fakeshield}, although it is trained on both GAN-generated and diffusion-edited images, it performs well on GAN-generated images but suffers a significant accuracy drop on diffusion-model-generated images, achieving only 52\%. SIDA~\cite{huang2025sida} LEGION~\cite{kang2025legion} and FakeShield~\cite{xu2024fakeshield} all yield detection accuracies close to random guessing, with accuracies on real images of 9.114\%, 33.60\%, and 0\%, respectively. For LEGION~\cite{kang2025legion}, detection is decoupled from localization and explanation, and only the weights of the localization and explanation modules are publicly available. During inference, it explains almost all images as fake, revealing a clear inconsistency between the classification head and textual outputs. For SIDA~\cite{huang2025sida} and FakeShield~\cite{xu2024fakeshield}, both methods integrate SAM~\cite{kirillov2023segment_anything} into the MLLM framework for text-guided forgery localization. However, SAM~\cite{kirillov2023segment_anything} is pretrained to segment all entities in an image rather than forged regions. Without explicit constraints from the detection objective, the localization branch tends to assign manipulation masks to authentic regions, leading to high false positive rates.

In contrast, our method demonstrates strong generalization across diverse generative models and achieves robust performance on both real and fake images, while maintaining consistency between detection results and reasoning outputs. Although trained solely on Stable Diffusion images, it generalizes well to images generated by the latest diffusion models and unseen GANs, exhibiting strong in-domain and cross-domain robustness.

\subsubsection{Robustness on Forgery Detection Task}

\begin{figure}[t]
    \centering
    \includegraphics[width=0.85\columnwidth]{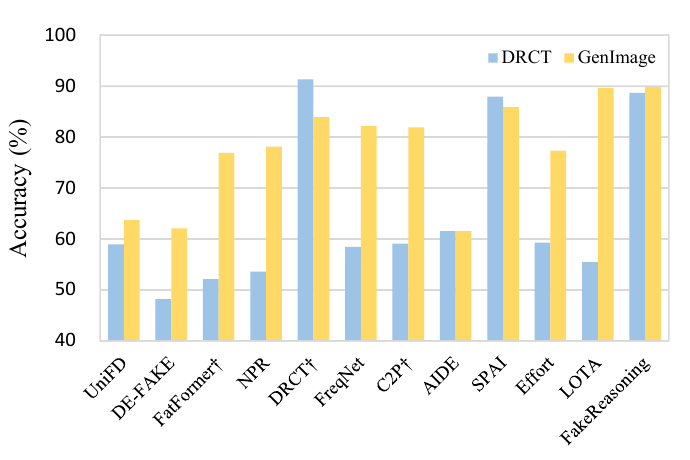}
    \caption{Detection performance on the DRCT-2M and GenImage dataset. ${\dagger}$ denotes using official weights and the others are trained on MMFR-Dataset.}
    \label{fig:more_dataset}
\end{figure}

\begin{table}[t]
\centering
\caption{Evaluation of the FDR-Task on the LOKI benchmark.}
\renewcommand{\arraystretch}{1.2}
\label{tab:loki}
\begin{tabular}{lccccc}
\toprule
\multirow{2}{*}{Method} & \multicolumn{2}{c}{Detection} & \multicolumn{3}{c}{Reasoning} \\ 
\cmidrule(lr){2-3} \cmidrule(lr){4-6}
& ACC & Fail $\downarrow$ & BL-1 & R-L & CSS \\ 
\midrule
FakeShield \cite{xu2024fakeshield}    & 49.50 & 2.00 & 0.13 & 0.12 & 0.51 \\
SIDA \cite{huang2025sida}             & 48.92 & 7.00 & 0.20 & 0.20 & 0.59 \\
LEGION \cite{kang2025legion}          & 50.00 & 1.00 & 0.18 & 0.20 & 0.61 \\
\midrule
FakeReasoning                          & \textbf{86.73} & \textbf{0.00} & \textbf{0.19} & \textbf{0.21} & \textbf{0.63} \\
\bottomrule
\end{tabular}
\end{table}

We further conduct robustness experiment on the evaluation sets to assess the stability of forgery detectors under real-world image degradations. Specifically, we examine three post-processing operations: JPEG compression, resizing, and Gaussian noise. Following the experiment setting of DRCT \cite{chendrct}, we use JPEG quality factors {100, 90, 80, 70, 60}, resize ratios {0.6, 0.8, 1.0, 1.2, 1.4}, and Gaussian noise levels with standard deviations {0, 3, 6, 9, 12}.
Also, we select UniFD \cite{ojha2023towards} and NPR \cite{tan2024rethinking} as representative high-level semantic and low-level artifact baselines, respectively, and compare them with the proposed FakeReasoning. All methods are trained on the MMFR-Dataset, and the results are shown in Figure \ref{fig:robust_exp}.

The results show that NPR \cite{tan2024rethinking} is extremely vulnerable to all three types of image degradation, with its accuracy dropping rapidly to near-random levels even under mild perturbations. This is because the low-level artifacts, such as upsampling traces, frequency cues, noise residuals, and reconstruction artifacts, are inherently fragile and easily disrupted by post-processing operations.
In contrast, high-level semantic information is more stable. UniFD \cite{ojha2023towards} employs a frozen CLIP visual encoder with a linear classifier, and its decisions rely primarily on global semantic distributions. As a result, it exhibits strong robustness across all perturbation types, consistent with prior findings on the robustness of CLIP.

Under the same perturbation conditions, FakeReasoning consistently achieves the highest detection accuracy and maintains superior robustness across different types and severities of image degradation. This demonstrates that FakeReasoning not only inherits the stable semantic space of CLIP but also incorporates fine-grained forgery-aware cues, resulting in enhanced overall detection performance. 

\subsection{Evaluation on Existing Benchmarks}

\begin{figure}[t]
    \centering
    \includegraphics[width=0.98\columnwidth]{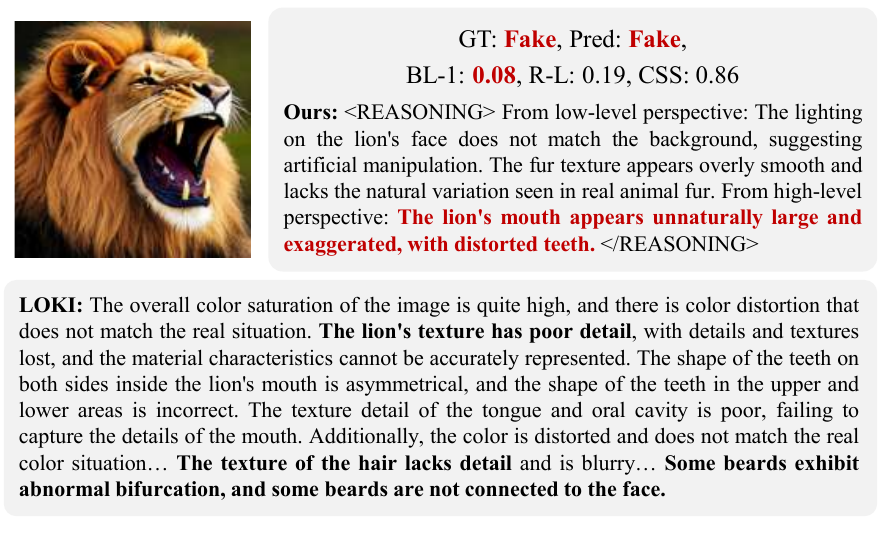}
    \caption{Visualization result on LOKI benchmark with low BLEU-1 score.}
    \label{fig:loki_vis}
\end{figure}

In this section, we choose LOKI \cite{ye2024loki} to evaluate both forgery detection and reasoning performance. Specifically, we use all the images in LOKI to evaluate forgery detection task and use ``open ended VQA'' to evaluate forgery reasoning task. Also, we choose DRCT-2M \cite{chendrct} and GenImage \cite{zhu2023genimage} to evaluate forgery detection performance.

\subsubsection{Evaluation on DRCT-2M and GenImage Dataset}
We further evaluate detection performance on both DRCT-2M \cite{chendrct} and GenImage \cite{zhu2023genimage}. Note that FakeReasoning is an MLLM-based framework, which cannot be trained on these solely image datasets. To ensure fair comparison, we train all methods on the MMFR-Dataset training set and evaluate on the DRCT-2M \cite{chendrct} and GenImage \cite{zhu2023genimage} test sets. In addition, since some methods do not provide training code (FatFormer \cite{liu2024visual}, C2P \cite{tan2025c2p}) or require extra reconstruction data (DRCT \cite{chendrct}), we use their official weights for evaluation. The results are shown in the Figure \ref{fig:more_dataset}. 
Almost all methods achieve higher performance on GenImage \cite{zhu2023genimage} than on DRCT-2M \cite{chendrct}. This gap can be attributed to two factors. First, DRCT-2M \cite{chendrct} includes image-editing forgeries in which the manipulated regions are relatively small, making them more difficult to detect. Second, DRCT-2M \cite{chendrct} contains JPEG-compressed images, which pose additional challenges for methods that rely on low-level artifacts. Among all methods, DRCT \cite{chendrct}, SPAI \cite{karageorgiou2025any}, and FakeReasoning demonstrate consistent performance across the two datasets, indicating stronger generalization. Given that the official weight of DRCT \cite{chendrct} are trained on DRCT-2M \cite{chendrct}, and thus its slightly higher performance than our method on DRCT-2M \cite{chendrct} is expected. Apart from this, FakeReasoning achieves the best performance on both datasets, demonstrating its strong generalization across different types of forgeries and generative paradigms.

\begin{figure*}
    \centering
    \includegraphics[width=\textwidth]{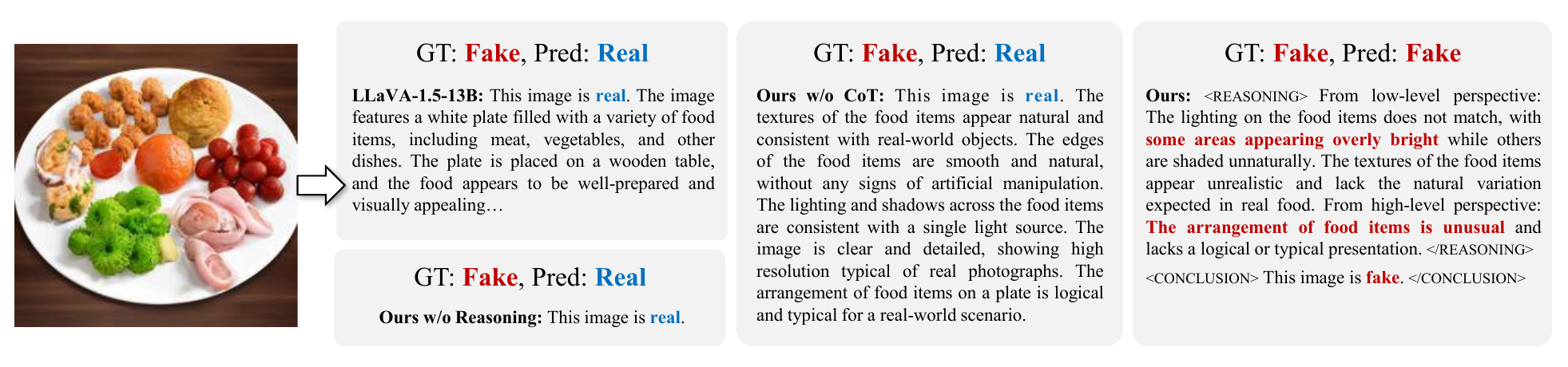}
    \caption{Visual examples in the ablation study on forgery reasoning task. Only reasoning and conclusion stages of our method is presented. With the forgery reasoning chain-of-thought, our method achieves accurate detection through structured and hierarchical reasoning.}
    \label{fig:forgery reasoning}
\end{figure*}

\begin{table}[t]
\caption{Ablation study on forgery reasoning task and CPM.}
\renewcommand{\arraystretch}{1.2} 
\centering
\begin{tabular}{l@{\hskip 6pt}c@{\hskip 6pt}c@{\hskip 6pt}c@{\hskip 6pt}c@{\hskip 6pt}c@{\hskip 6pt}c}
\toprule
\multirow{2}{*}{Variants} & \multicolumn{3}{c}{Detection} & \multicolumn{3}{c}{Reasoning}\\ \cmidrule(lr){2-4} \cmidrule(lr){5-7} 
 & \multicolumn{1}{c}{DM} & \multicolumn{1}{c}{GAN} & \multicolumn{1}{c}{AVG} & \multicolumn{1}{c}{BL-1} & \multicolumn{1}{c}{R-L} & \multicolumn{1}{c}{CSS} \\ 
 \midrule
LLaVA-1.5-13B \cite{liu2023improvedllava}       & 73.38  & 70.85  & 72.87  & 0.23   & 0.20   & 0.54 \\
Ours w/o Reasoning    & 90.25  & 93.50  & 90.90  & 0.0004 & 0.05  & 0.42 \\
Ours w/o CoT          & 92.63  & 94.50  & 93.00  & 0.36   & 0.29  & 0.76 \\
 \midrule
Ours w/o CPM       & 92.57 & 95.60 & 93.18 & 0.36 & 0.29 & 0.76 \\
 \midrule
Ours                  & 94.08  & 94.00  & 94.06  & 0.37   & 0.29   & 0.76 \\
\bottomrule
\end{tabular}
\label{tab:Ablation on Forgery Reasoning Task and cpm}
\end{table}

\begin{table}[t]
\centering
\caption{Ablation study on the Visual Branch. CA denotes Cross Attention Mechanism. Bias denotes Attention Bias. Q denotes Query and KV denotes Key/Value.}
\label{tab:Ablation on low-level vision}
\renewcommand{\arraystretch}{1.2} 
\begin{tabular}{c@{\hskip 5pt}c@{\hskip 5pt}c@{\hskip 5pt}c@{\hskip 9pt}c@{\hskip 9pt}c@{\hskip 9pt}c@{\hskip 9pt}c@{\hskip 9pt}c@{\hskip 9pt}c}
\toprule
\multirow{2}{*}{CLIP} & \multirow{2}{*}{DINO} & \multirow{2}{*}{CA} & \multirow{2}{*}{Bias} 
& \multicolumn{3}{c}{Detection} & \multicolumn{3}{c}{Reasoning} \\
\cmidrule(lr){5-7} \cmidrule(lr){8-10}
 & & & & DM & GAN & AVG & BL-1 & R-L & CSS \\
\midrule
\checkmark & × & × & × & 90.33 	& 87.85 & 89.83  & 0.36 & 0.29 & 0.76 \\
\checkmark & \checkmark & × & × & 93.75 & 89.00 & 92.80 & 0.36 & 0.29 & 0.76 \\
\checkmark & \checkmark & \checkmark & × & 90.38 & 96.00 & 91.50 & 0.36 & 0.29 & 0.76 \\
\midrule
KV & Q & \checkmark & \checkmark & 89.11 & 96.23 & 90.53 & 0.35 & 0.29 & 0.74 \\

\midrule
Q  & KV & \checkmark & \checkmark & 94.08 & 94.00 & 94.06 & 0.37 & 0.29 & 0.76 \\
\bottomrule
\end{tabular}
\end{table}

\subsubsection{Evaluation on LOKI Benchmark}
As shown in Table~\ref{tab:loki}, existing MLLM-based detectors achieve only near–random-guessing detection accuracy on the LOKI benchmark, primarily due to their high false positive rates on real images. In contrast, our method demonstrates superior detection and reasoning capabilities compared to MLLM-based detectors. 

Besides, we observe that our method exhibits a decline in BLEU-1 and ROUGE-L scores on the LOKI benchmark compared to the MMFR-Dataset. Through further analysis, we find that the 229 reasoning samples in the open-ended VQA comprises certain satellite and document images. These images differ significantly from natural images in reasoning patterns, resulting in reduced lexical coverage and disrupted word order alignment. 
Figure \ref{fig:loki_vis} shows a representative case with correct prediction despite a low BLEU-1 score. Notably, the LOKI's annotation contains semantic redundancy (``lion's texture has poor detail'' and ``texture of the hair lacks detail'') as well as fine-grained hallucination (``some beards exhibit abnormal bifurcation...not connected to the face''). In contrast, our model performs reliable reasoning from both low-level and high-level perspectives and successfully identifies a key cue, that the abnormal size and distortion of the lion’s mouth and teeth, which is not included within the ground truth annotation.

\subsection{Ablation Study}

\subsubsection{Ablation Study on Forgery Reasoning Task} 

In this section, we try to validate that the FDR-Task, which couples visual reasoning with forgery detection, not only provides more explainable detection results, but also fully exploits MLLMs for image forensics. We design the following variants: 1) \textbf{Ours w/o Reasoning:} the variant takes ``This image is real.'' or ``This image is fake.'' as the ground truth output; 2) \textbf{Ours w/o CoT:} the variant takes original interpretations of GPT-4o as ground truth output. The performance of these variants is shown in Figure \ref{fig:forgery reasoning} and Table \ref{tab:Ablation on Forgery Reasoning Task and cpm}. 

Fine-tuning MLLMs with simple binary answers could enhance their detection performance. However, this variant significantly constrains MLLMs' reasoning capability, as it is difficult to align complex visual content with a binary label sentence. Conducting the reasoning task alongside detection not only enhances the quality of reasoning but also improves detection accuracy. Meanwhile, integrating \textit{Forgery Reasoning CoT} further boosts both detection and reasoning performance. This structured process facilitates better alignment between visual and linguistic modalities, while hierarchical reasoning effectively leverages both high-level and low-level visual clues from the visual branch.

\subsubsection{Ablation Study on the Visual Branch}

In this work, we introduce a dual-branch visual encoder to enhance the fine-grained perceptual capability of MLLMs. To effectively integrate high-level and low-level visual clues, we propose a Forgery-Aware Feature Fusion module that enables discriminative feature learning for forgery detection. To validate the effectiveness of each component, we design several ablation variants. As shown in in Table \ref{tab:Ablation on low-level vision}, \textbf{CLIP} denotes using only the CLIP vision encoder; \textbf{CLIP+DINO} represents a simple interleaving of DINO and CLIP embeddings following \cite{tong2024eyes}; \textbf{CLIP+DINO+CA} applies a cross-attention mechanism to fuse DINO and CLIP features; and \textbf{CLIP+DINO+CA+Bias} further incorporates DINO's attention maps as the attention bias to guide the cross-attention integration. Since DINO is pretrained without textual supervision, using DINO alone struggles to align with LLMs, and thus a DINO-only variant is not included. According to the results of \textbf{CLIP+DINO}, introducing the DINO encoder yields significant improvements in forgery detection. This confirms the feature-level complementarity between DINO and CLIP in MLLM-based forensics. Although \textbf{CLIP+DINO+CA} shows a slight accuracy drop, \textbf{CLIP+DINO+CA+Bias} achieves the best detection and reasoning performance by introducing DINO’s attention maps as forgery-aware priors. 

To verify the effectiveness of this directional choice, we further conduct an ablation study comparing two configurations: CLIP→DINO (Q→KV) and DINO→CLIP (Q→KV), with results shown in Table \ref{tab:Ablation on low-level vision}. When CLIP serves as the Query, the variant achieves superior performance in both forgery detection and reasoning tasks, with an average detection accuracy improvement of approximately 3.5\%. The CLIP→DINO direction enables semantically guided attention to focus on forgery-related regions and capture semantic inconsistencies more effectively. On the contrary, using DINO as the Query leads to low-level feature dominance, which weakens semantic focus and degrades cross-modal alignment, resulting in reduced detection and reasoning performance.

\begin{table}[t]
\centering
\begin{minipage}{0.48\linewidth}
\centering
\caption{Ablation study on the layers.}
\label{tab:ablation_layer}
\renewcommand{\arraystretch}{1.2}
\begin{tabular}{cccc}
\toprule
Layer & DM & GAN & ACC \\
\midrule
1st  & 91.63 & 95.00 & 92.30 \\
6th  & 91.50 & 95.50 & 92.30 \\
12th & 88.75 & 96.00 & 90.20 \\
18th & 92.75 & 95.00 & 93.20 \\
24th & 94.08 & 94.00 & 94.06 \\
\bottomrule
\end{tabular}
\end{minipage}
\hspace{1pt} 
\begin{minipage}{0.48\linewidth}
\centering
\caption{Ablation study on $\tau$.}
\label{tab:ablation_tao}
\renewcommand{\arraystretch}{1.2}
\begin{tabular}{cccc}
\toprule
$\tau$ & DM & GAN & ACC \\
\midrule
1  & 92.99 & 93.00 & 92.99 \\
5  & 92.04 & 94.50 & 92.53 \\
10 & 94.08 & 94.00 & 94.06 \\
15 & 92.42 & 93.00 & 92.54 \\
20 & 92.79 & 95.00 & 93.23 \\
\bottomrule
\end{tabular}
\end{minipage}
\end{table}

\begin{table}
\caption{Ablation study on initialization, dataset and framework.}
\renewcommand{\arraystretch}{1.2} 
\centering
\begin{tabular}{l@{\hskip 6pt}c@{\hskip 6pt}c@{\hskip 6pt}c@{\hskip 6pt}c@{\hskip 6pt}c@{\hskip 6pt}c}
\toprule
\multirow{2}{*}{Variants} & \multicolumn{3}{c}{Detection} & \multicolumn{3}{c}{Reasoning}\\ \cmidrule(lr){2-4} \cmidrule(lr){5-7} 
 & \multicolumn{1}{c}{DM} & \multicolumn{1}{c}{GAN} & \multicolumn{1}{c}{AVG} & \multicolumn{1}{c}{BL-1} & \multicolumn{1}{c}{R-L} & \multicolumn{1}{c}{CSS} \\ 
 \midrule
LLaVA-1.5-13B  \cite{liu2023improvedllava} & 73.38  & 70.85  & 72.87  & 0.23   & 0.20   & 0.54 \\ 
+ Dataset & 92.87   & 92.95  & 92.88  & 0.36  & 0.30  & 0.77  \\
+ Framework & 92.25 & 95.75 & 92.95 & 0.37 & 0.30 & 0.77 \\
 \midrule
LLaVA-MoF  \cite{tong2024eyes} & 72.90 & 67.19 & 71.76 & 0.24 & 0.20 & 0.54 \\
+ Dataset & 92.86 & 91.50 & 92.59 & 0.36 & 0.30 & 0.77 \\
+ Framework & 94.08 & 94.00 & 94.06 & 0.37 & 0.29 & 0.76 \\
\bottomrule
\end{tabular}
\label{tab:Ablation on initialization}
\end{table}

To further analysis the accuracy gain by the attention bias, we conduct an ablation study on the layer choice of DINO’s attention maps. Specifically, we extract attention maps from the 1st, 6th, 12th, 18th, and 24th layers of the pretrained DINO encoder and use them as the attention bias in FAFF module. The result are shown in Table \ref{tab:ablation_layer}. Using the attention maps from the 24th layer achieves the highest average accuracy. These results indicate that leveraging high-level attention maps from DINO as the attention bias produces more discriminative fused features. We attribute this to two key factors. First, higher-layer attention maps tend to focus on semantically meaningful and context-aware regions, which can better highlight potential forgery areas and entities. Second, since CLIP and DINO features are first projected into semantic space via MLP layers, high-level attention aligns more effectively with the semantics of these adapted features.

\subsubsection{Ablation Study on CPM}

\begin{figure}[t]
\centering
\begin{minipage}{0.45\linewidth}
    \centering
    \includegraphics[width=\linewidth]{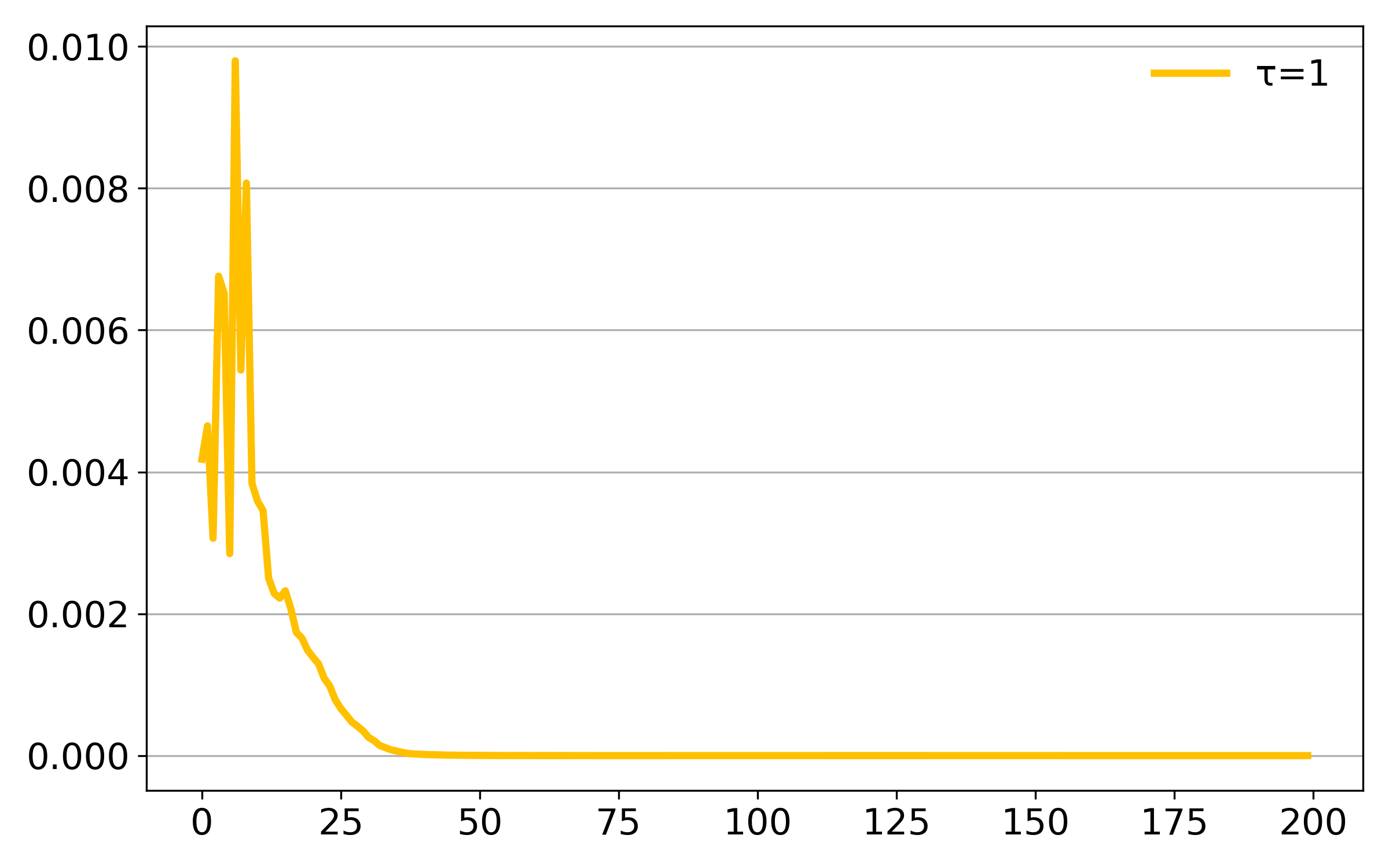}
    \vspace{2pt}
    (a)
\end{minipage}
\hspace{1pt}
\begin{minipage}{0.45\linewidth}
    \centering
    \includegraphics[width=\linewidth]{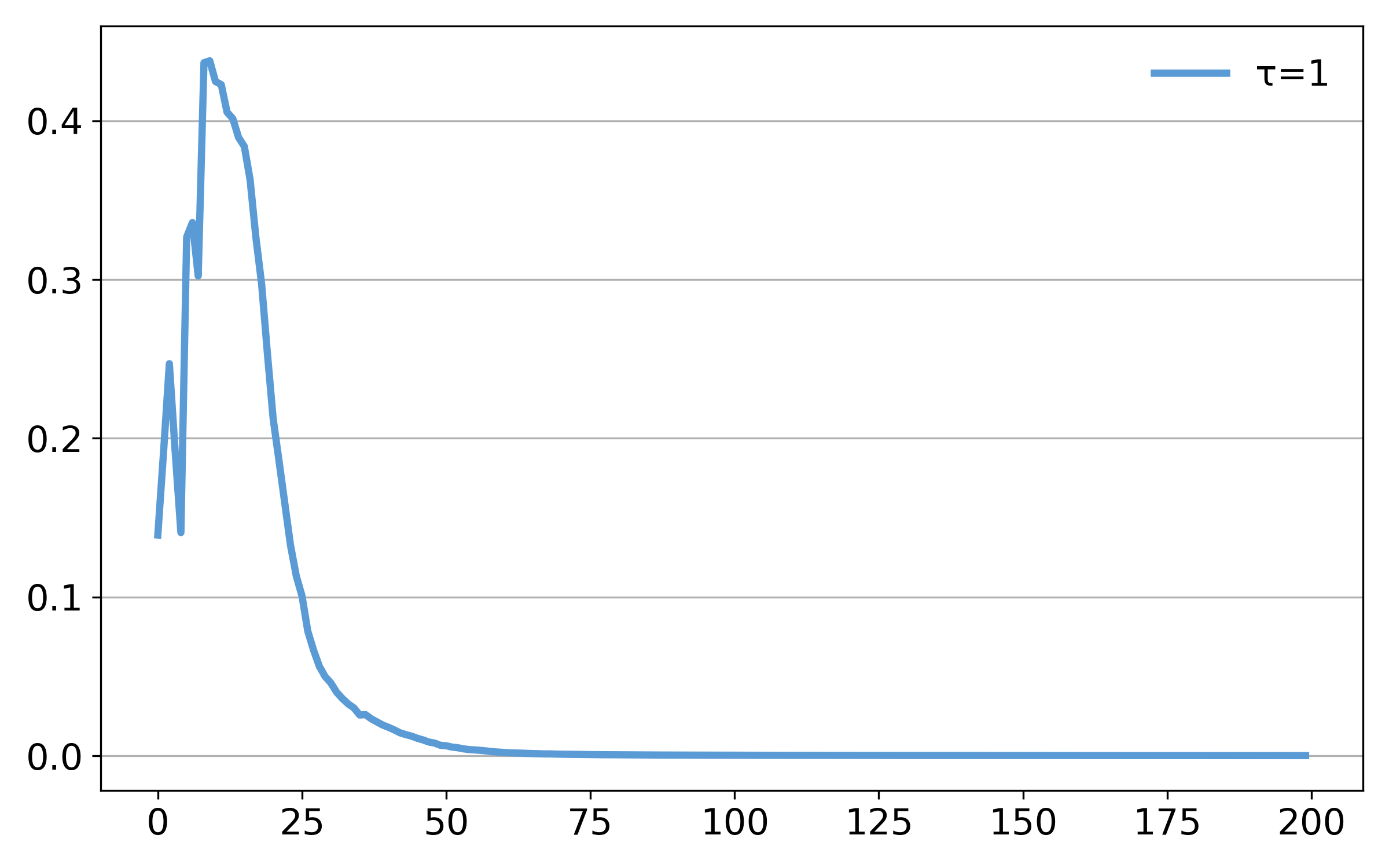}
    \vspace{2pt}
    (b)
\end{minipage}
\caption{(a) $\mathcal{L}_{CE}$ curve at $\tau = 1$. (b) $\mathcal{L}_{CE}$ curve at $\tau = 10$.}
\label{fig:ce loss}
\end{figure}

In our work, we propose the Classification Probability Mapper to better couple forgery detection with language modeling. As illustrated in Table~\ref{tab:Ablation on Forgery Reasoning Task and cpm}, \textbf{Ours w/o CPM} denotes using only the language modeling objective without any classification guidance. According to the results, the CPM module significantly improves the detection accuracy. Through probability mapping, the CPM module unifies the optimization objectives and enables coupling between forgery detection and language modeling. As a result, it not only mitigates the conflict between tasks but also achieves accuracy gain in forgery detection.

To justify the selection of this hyperparameter, we first evaluate the detection accuracy under $\tau \in \{1, 5, 10, 15, 20\}$. As shown in Table \ref{tab:ablation_tao}, $\tau = 10$ achieves the highest accuracy. Furthermore, we plot the $\mathcal{L}_{CE}$ curve of first 200 steps under $\tau = 1$ and $\tau = 10$ in Figure \ref{fig:ce loss}. When $\tau = 1$, the $\mathcal{L}_{CE}$ remains extremely small and quickly converges to zero, suggesting that the optimization is dominated by the $\mathcal{L}_{LM}$. In contrast, when $\tau = 10$, the $\mathcal{L}_{CE}$ starts from a reasonable scale and decreases smoothly. Therefore, $\tau = 10$ is adopted as the default setting in the proposed FakeReasoning.

As $\mathcal{L}_{CE}$ of unmatched batches is set to 0, Figure \ref{fig:ce loss}(b) illustrates a slight increase within the first ten steps. After that, our model rapidly learns to output standardized conclusion templates and the loss peaks before steadily converging toward zero. This behavior demonstrates that the fixed template matching mechanism stabilizes quickly after the warm-up phase, and the overall convergence curve aligns well with expectations.

\subsubsection{Ablation Study on Initialization}

To clarify the effect of pre-trained initialization, we further conduct ablation study comparing different initialization and fine-tuning configuration.
As shown in Table~\ref{tab:Ablation on initialization}, replacing the initialization from LLaVA-1.5 with that from LLaVA-MoF \cite{tong2024eyes} leads to only a marginal improvement in average detection accuracy (from 92.95\% to 94.06\%) and comparable reasoning performance. The initialization from LLaVA-MoF is primarily adopted to inherit the DINO adapter weights, which provide a stronger multi-modal alignment prior and facilitate more stable convergence. However, this initialization is not the main factor contributing to the superior performance of FakeReasoning. In contrast, as shown in Table~\ref{tab:Ablation on initialization}, fine-tuning on the MMFR-Dataset and adopting the proposed framework jointly yield substantial gains in both detection and reasoning performance. This demonstrates that the main performance improvement arises from our fine-tuning framework and dataset rather than the pre-trained initialization.

\subsection{Visualization}

\begin{figure*}
    \centering
    \includegraphics[width=\textwidth]{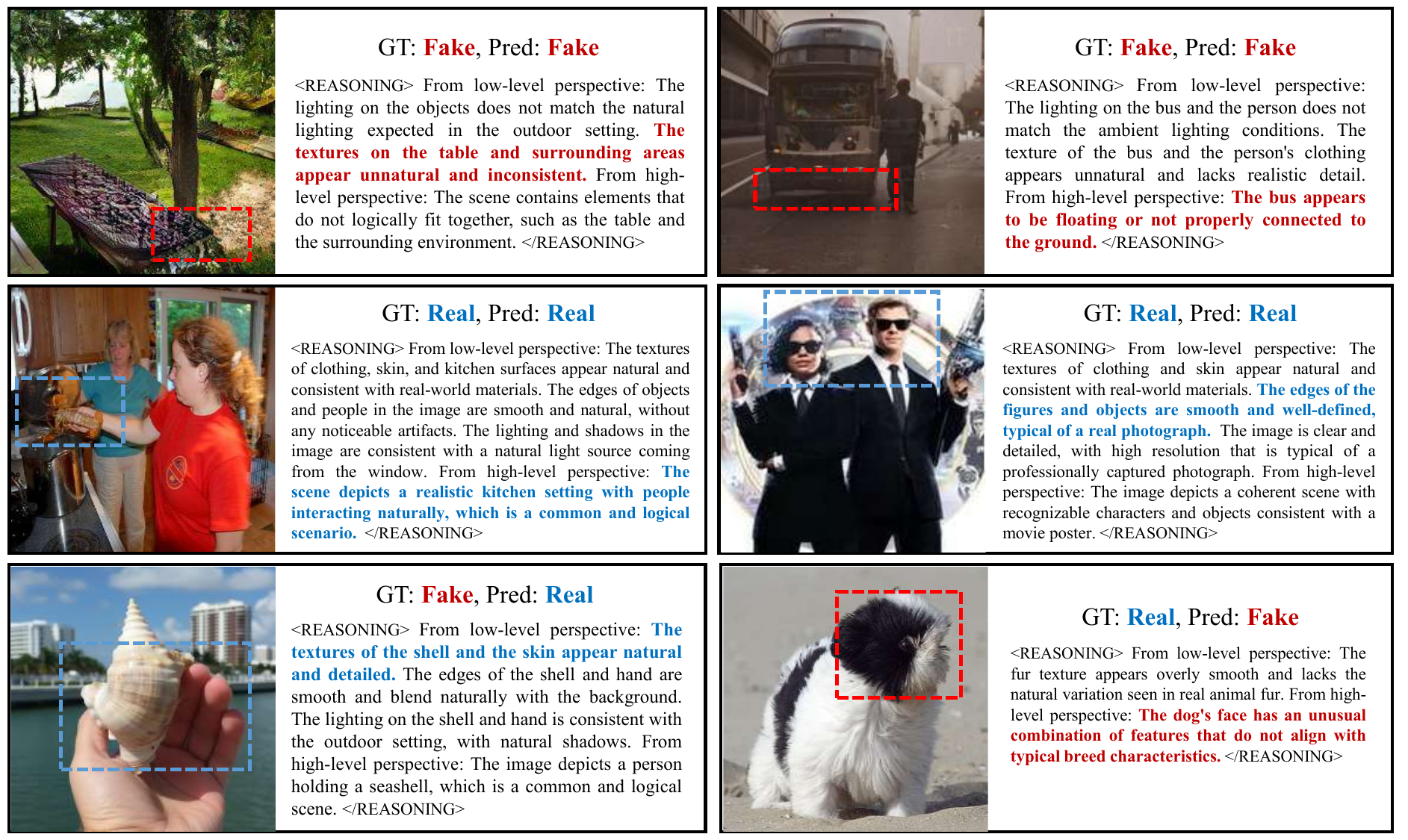}
    \caption{Visualization results of FakeReasoning. Discriminate reasoning clues are bolded and corresponding regions are highlighted with dashed boxes.}
    \label{fig:visualization}
\end{figure*}

We provide the forgery detection and reasoning results of the proposed FakeReasoning in Figure \ref{fig:visualization}. Discriminate reasoning clues are bolded and corresponding regions in are highlighted within images. 

As shown in the first two rows, FakeReasoning provides accurate detection results as well as reliable forgery reasoning from both high-level semantics and low-level artifacts. The third row shows representative failure cases. First, we analyze several fake images that are misclassified as real and find that these images often exhibit high visual fidelity. For example, in the first image of the third row, the shell texture is finely rendered, and both the hand structure and fingerprint details appear highly realistic, making it nearly impossible to distinguish even with the naked eyes. As generative models continue to advance, producing perfect fake images is easier than ever. Recent studies have demonstrated that MLLMs can perceive modalities beyond human vision, such as infrared and depth images. Inspired by these advances, our future work will investigate whether MLLMs can identify imperceptible artifacts and how to aligned artifacts with natural language.

Besides, we also analyze real images that are misclassified as fake, which can be broadly categorized into three types: 1) unconventional structures: as shown in the second image in the third row, the dog is positioned in a side-facing pose, revealing only one eye, with its dense fur obscuring the facial features, which deviates from the typical morphology of common dogs; 2) prominent foregrounds: typically captured with telephoto lenses that blur the background heavily, creating a visual pattern similar to generators' lack of background realism, thus leading to misclassification; 3) abnormal texture: post-processing operations like sharpening or contrast enhancement cause texture deviations from typical distributions, interfering with the predictions. To address these issues, our future work will concentrate on expand data diversity, including complex distributions and post-processing operations.

\subsection{More Analysis}
\subsubsection{Computational Cost Comparison with MLLM-based Detectors}
We compare the proposed FakeReasoning with representative MLLM-based detectors on the computational cost, reporting trainable parameters, inference latency, and FLOPs. For methods employing multi-stage training, trainable parameters of each stage are reported separately with a slash. All the inference experiments are conducted on a single NVIDIA GeForce RTX 4090 GPU with a batch size of 1 and an input resolution of 1024×1024. Inference latency is reported as the average processing time over 100 images. As the output length of MLLMs varies across methods and images, FLOPs are measured for the prefill phase \cite{agrawal2023sarathi} with generation length fixed to one token. 
Results are shown in Table \ref{tab:cost}. FakeReasoning achieves the lowest inference latency among all methods, while maintaining competitive FLOPS relative to FakeShield \cite{xu2024fakeshield} and LEGION \cite{kang2025legion}. Notably, FakeReasoning is trained in an end-to-end manner, rather than following the two-stage training pipelines of other MLLM-based detectors, which effectively reduces the model complexity.

\begin{table}[t]
\renewcommand{\arraystretch}{1.2} 
\centering
\caption{Computational cost comparison with MLLM-based detectors.}
\label{tab:cost}
\begin{tabular}{lccc}
\toprule
Method & Parameters & Latency & FLOPs \\ 
\midrule
FakeShield \cite{xu2024fakeshield}   & 532M / 587M & 12470 ms & 19.19T \\
SIDA \cite{huang2025sida}        & 216M / 351M & 36820 ms & 8.59T  \\
LEGION \cite{kang2025legion}      & 587M / 2M   & 17870 ms & 16.45T \\
\midrule
FakeReasoning & 878M        & 9820 ms  & 17.97T \\
\bottomrule
\end{tabular}
\end{table}

\subsubsection{Consistency Analysis on Forgery Detection Task} We perform consistency analysis on the detection results to assess potential inconsistencies in MLLM-based detectors. Specifically, we randomly sample 100 images from each evaluation set and run 100 inference rounds on these samples, with the initial temperature set to 0.2. The majority prediction is taken as the baseline, and the proportion of minority predictions is calculated to quantify inconsistency. The average inconsistency of forgery detection is 0.80\%.

Temperature regulates randomness in text generation. Higher temperatures lead to more random and diverse outputs, while lower temperatures yield more deterministic and conservative results. We further evaluate the impact of temperature on inconsistency by setting it to 0.1, 0.2, 0.3 and 0.4. As the temperature increases, the inconsistency becomes more pronounced. The average inconsistency remains below 1\% at temperatures of 0.1 and 0.2. 

\subsubsection{Human Evaluation on Forgery Reasoning Task} We conduct human evaluation on the reasoning outputs to assess potential hallucinations in MLLM-based detectors. Specifically, we randomly sample 100 images from each evaluation set, where both the ground-truth and predicted labels are fake. Each sentence in the forgery reasoning is scored using a 5-grade marking method by human experts, with 5 indicating highly rational and 1 indicating highly irrational. The average score is then computed for each evaluation set. The average rationality score for forgery reasoning is 4.19. And the results reveal a negative correlation between the rationality and the generative quality. When the generative quality is low, FakeReasoning provides detailed reasoning based on visual contents. As the generative quality increases, FakeReasoning tends to output more stereotypical reasoning results.

\section{Limitation}
As shown in the ablation study, the introduced modules in the visual branch have little impact on the metrics of the reasoning task. 
The phenomenon may be attributed from two perspectives. First, the reasoning process of current MLLMs is language-dominated, where visual embeddings mainly serve as auxiliary inputs rather than directly participating in the reasoning chain. In FakeReasoning, the enhanced visual branch effectively improves fine-grained perception of forgery cues, which benefits the detection task. However, its influence on the language-dominated reasoning mechanism itself remains relatively limited. Second, the effects of supervised fine-tuning and evaluation metrics also contribute to this phenomenon. The task-oriented SFT on MMFR-Dataset constrains the output text space toward the annotation distribution. Meanwhile, the reasoning task is evaluated by metrics that primarily rely on textual patterns and emphasize semantic similarity rather than visual dependency. As a result, the improvements brought by the enhanced visual branch may not be directly reflected in current reasoning metrics.
Fully leveraging MLLMs for forgery reasoning, particularly the language branch, remains an open challenge. 
Based on the above analysis, in future work we aim to allow forgery-related visual cues to play a more direct role in the reasoning chain rather than merely serving as inputs. Inspired by the emerging “think-with-image’’ paradigm, we plan to further investigate the potential of visual information to enhance the interpretability, consistency, and rationality of the reasoning process. 
Furthermore, we plan to extend this approach to multi-modal forgery detection and improve robustness under real-world distribution shifts, thereby advancing more reliable verification systems for AI-generated content.

\section{Conclusion}

In this paper, we introduced the FDR-Task, a framework that unifies detection and reasoning for AI-generated images. To support this, we proposed the MMFR-Dataset, a large-scale dataset with hierarchical forgery reasoning annotations. Leveraging this dataset, we developed FakeReasoning, which enhances both generalization and interpretability by adapting MLLMs' visual branch and optimization to image forensics. Experiments on multiple generative models demonstrate that FakeReasoning consistently outperforms state-of-the-art methods in both detection and reasoning tasks.

\section*{Acknowledgments}
This work was supported by the National Natural Science Foundation of China (Grant 62406171, 62225601, U23B2052, 62406172), in part by the Beijing Natural Science Foundation Project No. L242025, in part by the Fundamental Research Funds for the Beijing University of Posts and Telecommunications under Grant 2025AI4S15, in part by the China Postdoctoral Science Foundation No. 2023M741964, and in part by the Postdoctoral Fellowship Program of CPSF No. GZC20240841.

\bibliographystyle{IEEEtran}
\bibliography{main}

\newpage

\section{Biography Section}
\vspace{-10pt}
\begin{IEEEbiography}[{\includegraphics[width=1in,height=1.25in,clip,keepaspectratio]{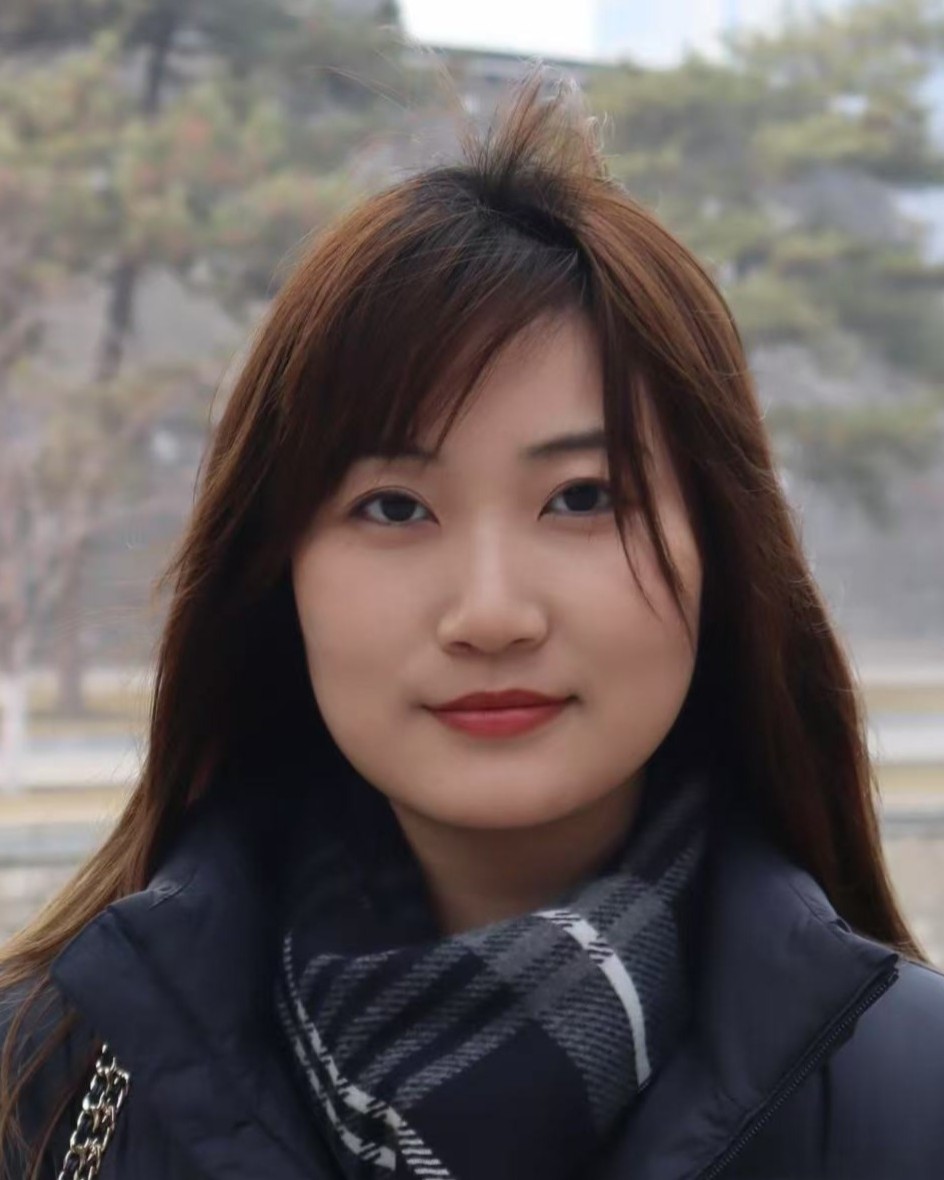}}]{Yueying Gao}
 received the M.Eng. degree in the Electronic Information engineering from the Communication University of China, in 2024. She is currently pursuing the Ph.D. degree with the Beijing University of Posts and Telecommunications. Her
 current research interests include the multi-media forensics and computer vision.
\end{IEEEbiography}

\vspace{-10pt}

\begin{IEEEbiography}[{\includegraphics[width=1in,height=1.25in,clip,keepaspectratio]{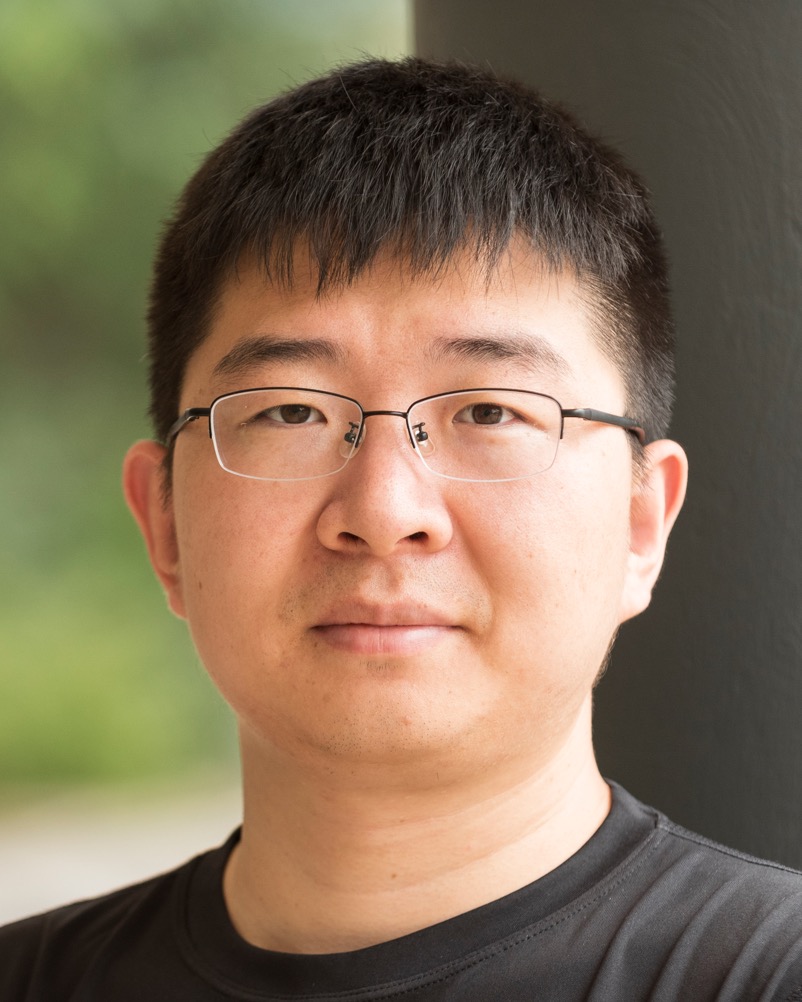}}]{Dongliang Chang}
received the Ph.D. degree in Information and Communication Engineering from Beijing University of Posts and Telecommunications, China, in 2023. He is currently a tenure-track assistant professor with the Beijing University of Posts and Telecommunications,  China, since 2025. From 2023 to 2025, he was a postdoctoral researcher fellow with the Department of Automation, Tsinghua University, China.  His research interests include deep learning and computer vision, particularly fine-grained visual understanding.
\end{IEEEbiography}

\vspace{-10pt}

\begin{IEEEbiography}[{\includegraphics[width=1in,height=1.25in,clip,keepaspectratio]{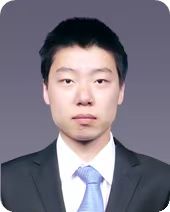}}]{Bingyao Yu}
Bingyao yu received the B.S. and Ph.D. degrees both in the Department of Automation, Tsinghua University, China, in 2018 and 2023. His current research interests include computer vision, deep learning, face forgery detection and face anti-spoofing. He serves as a regular reviewer member for a number of journals and conferences, e.g. TIP, ICML, ICLR, NeurIPS, ICCV, CVPR, and ECCV.
\end{IEEEbiography}

\vspace{-10pt}

\begin{IEEEbiography}[{\includegraphics[width=1in,height=1.25in,clip,keepaspectratio]{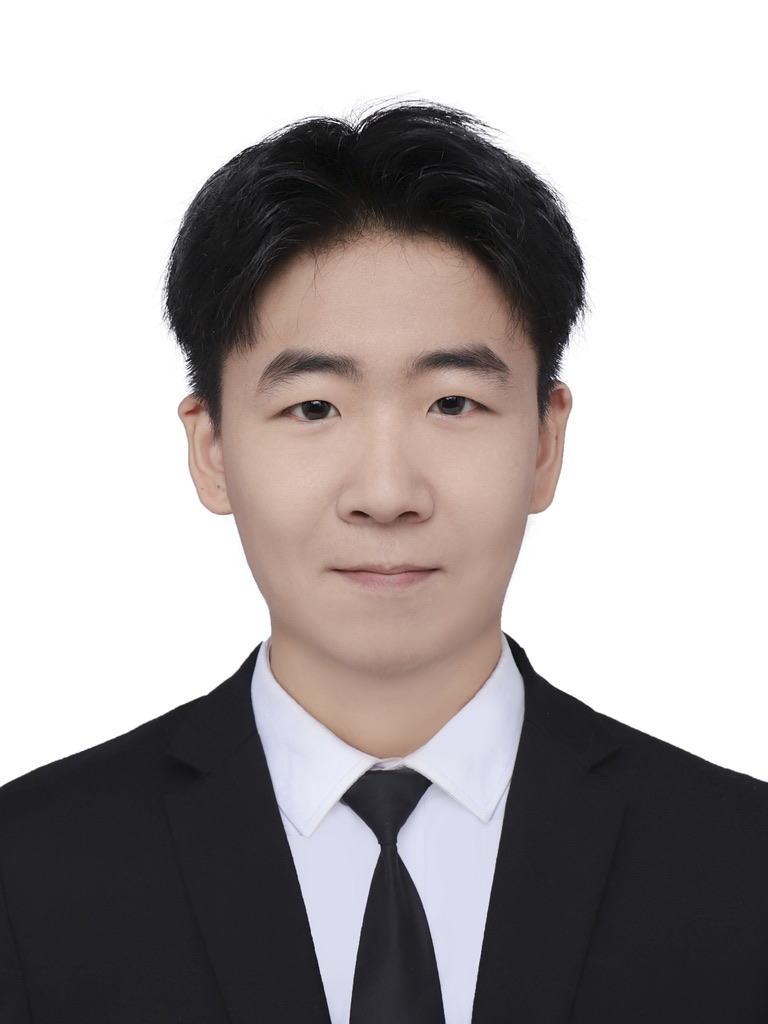}}]{Haotian Qin}
received the Bachelor of Science degree in Data Science and Big Data Technology from China University of Petroleum (East China), Qingdao, China, in 2020. He is currently a graduate student with the School of Artificial Intelligence, Beijing University of Posts and Telecommunications, Beijing, China. His research interests include AI-Generated image detection.
\end{IEEEbiography}

\vspace{-10pt}

\begin{IEEEbiography}[{\includegraphics[width=1in,height=1.25in,clip,keepaspectratio]{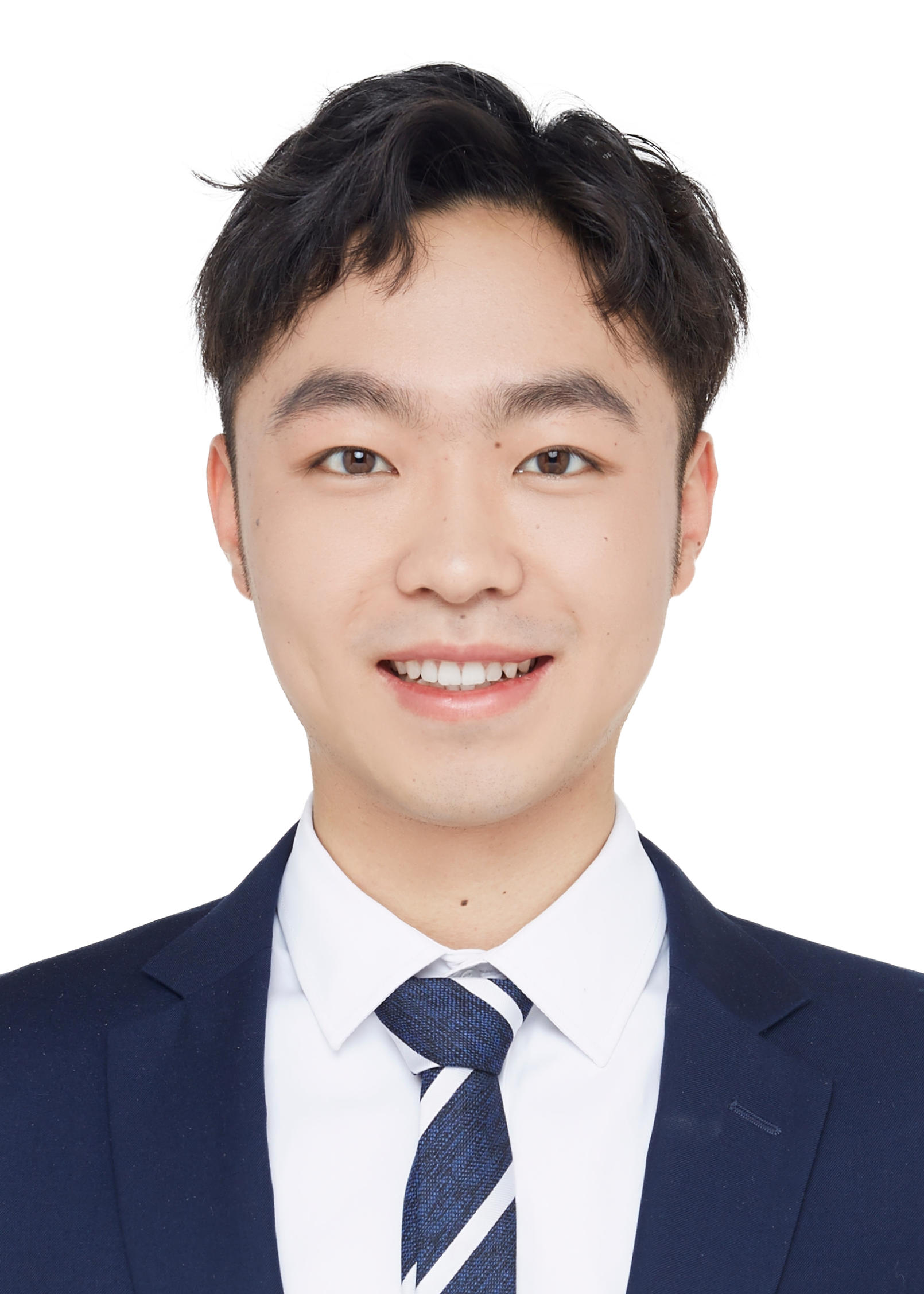}}]{Muxi Diao}
PhD student at the Pattern Recognition and Intelligent Systems Laboratory, Beijing University of Posts and Telecommunications. His main research direction is reinforcement learning for large multimodal foundation models, as well as large language model alignment and safety.
\end{IEEEbiography}

\vspace{-10pt}

\begin{IEEEbiography}[{\includegraphics[width=1in,height=1.25in,clip,keepaspectratio]{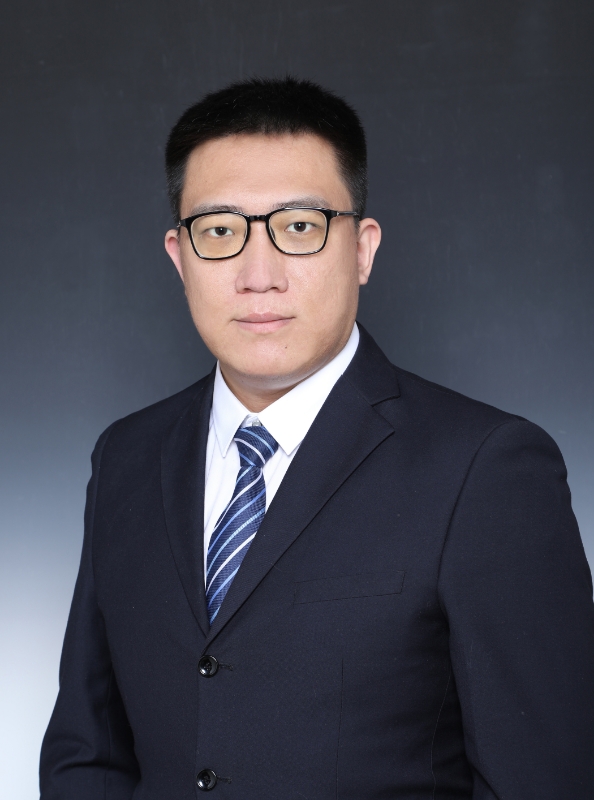}}]{Lei Chen}
(Member, IEEE) received the Ph.D. degree from Tianjin University, Tianjin, China. He is currently an Assistant Researcher with the Department of Automation, Tsinghua University, Beijing, China. His research interests include computer vision and pattern recognition, with a particular emphasis on human motion analysis and understanding. Dr. Chen has authored more than twenty papers in leading international journals and conferences, including IEEE Transactions on Circuits and Systems for Video Technology, Pattern Recognition, and the European Conference on Computer Vision. He regularly serves on the review committees of premier venues such as IEEE Transactions on Pattern Analysis and Machine Intelligence, IEEE Transactions on Circuits and Systems for Video Technology, IEEE/CVF Conference on Computer Vision and Pattern Recognition, and European Conference on Computer Vision.
\end{IEEEbiography}

\vspace{-10pt}

\begin{IEEEbiography}[{\includegraphics[width=1in,height=1.25in,clip,keepaspectratio]{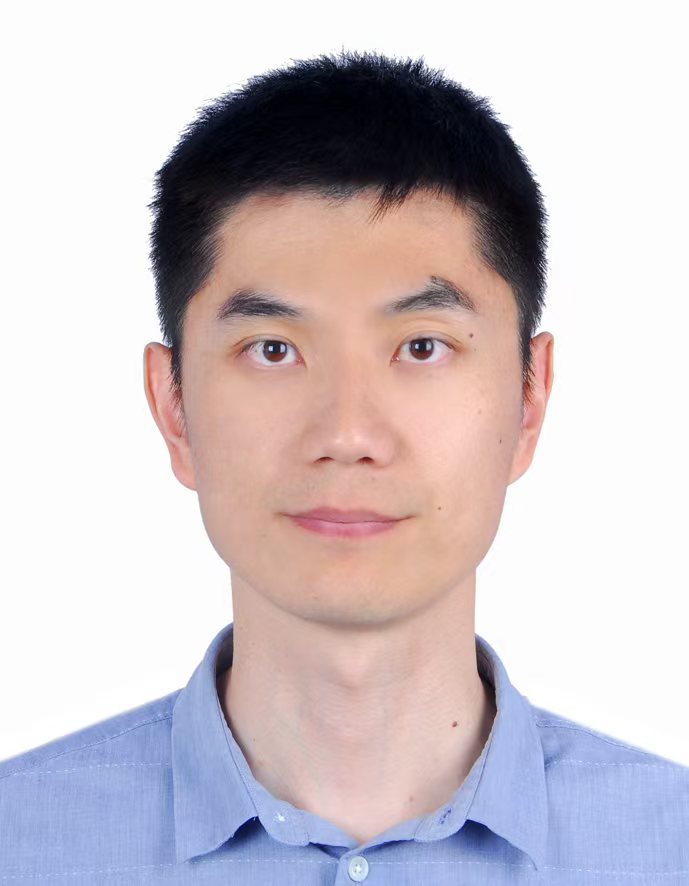}}]{Kongming Liang}
received the Bachelor’s degree from China University of Mining \& Technology, Beijing, China, in 2012; and the Ph.D. degree from Institute of Computing Technology, Chinese Academy of Sciences, Beijing, China, in 2018. He was a joint Ph.D. Student of machine learning group in Carleton University from Sep 2016 to Oct 2017 and a postdoc researcher in the Department of Computer Science at Peking University from Jan 2019 to Dec 2020. Currently, he is an associate professor of Beijing University of Posts and Telecommunications. His research interests cover computer vision and machine learning, especially visual concept understanding, multimodal large model optimization and evaluation, and medical image analysis.
\end{IEEEbiography}

\vspace{-10pt}

\begin{IEEEbiography}[{\includegraphics[width=1in,height=1.25in,clip,keepaspectratio]{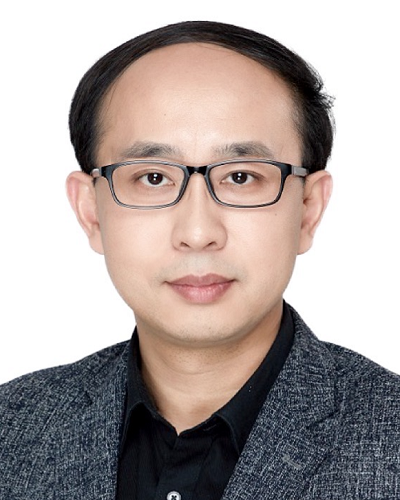}}]{Zhanyu Ma}
(Senior Member, IEEE) is currently a Professor at Beijing University of Posts and Telecommunications, Beijing, China, since 2019. He received the PhD degree in electrical engineering from KTH Royal Institute of Technology, Sweden, in 2011. From 2012 to 2013, he was a Postdoctoral Research Fellow with the School of Electrical Engineering, KTH. He has been an Associate Professor with
the Beijing University of Posts and Telecommunications, Beijing, China, from 2014 to 2019. His research interests include pattern recognition and machine learning fundamentals with a focus on applications in computer vision, multimedia signal processing.
\end{IEEEbiography}

\vfill

\end{document}